\documentclass[journal]{IEEEtran}

\usepackage{amssymb}
\usepackage{cite}
\usepackage{color}
\usepackage{amsthm}
\usepackage{booktabs}
\usepackage{graphicx}
\usepackage{subfigure}
\usepackage{epstopdf}
\usepackage{multirow}
\usepackage{algorithm}
\usepackage{algorithmic}
\ifCLASSINFOpdf
\else
\fi
\usepackage{amsmath}
\usepackage{algorithmic}
\usepackage{array}
\usepackage{url}
\usepackage{enumitem}

\graphicspath{ {./Figures/} }

\newtheorem{theorem}{Theorem}

\newtheorem{lemma}{Lemma}

\newtheorem{definition}{Definition}
\newtheorem{remark}{Remark}

\newcommand{\ba}{\begin{array}}
\newcommand{\ea}{\end{array}}
\newcommand{\be}{\begin{equation}}
\newcommand{\ee}{\end{equation}}

\newcommand{\RNum}[1]{\lowercase\expandafter{\romannumeral #1\relax}}
\newcommand{\RNumU}[1]{\uppercase\expandafter{\romannumeral #1\relax}}

\def\R{\mathbb{R}}

\def\A{{\bf A}}
\def\B{{\bf B}}

\def\H{{\bf H}}
\def\I{{\bf I}}

\def\U{{\bf U}}
\def\V{{\bf V}}
\def\W{{\bf W}}

\def\Y{{\bf Y}}

\def\CU{{\boldsymbol{\mathcal U}}}
\def\CV{{\boldsymbol{\mathcal V}}}
\def\CW{{\boldsymbol{\mathcal W}}}

\def\a{{\bf a}}
\def\am{{\phi_{\bf a}}}

\def\b{{\bf b}}
\def\e{{\bf e}}
\def\hd{{\rm hdmx}}

\def\u{{\bf u}}

\def\x{{\bf x}}
\def\y{{\bf y}}

\begin{document}
%
\title{\huge 0/1 Deep Neural Networks via Block Coordinate Descent
}

%
%
%

\author{Hui Zhang, Shenglong Zhou, Geoffrey Ye Li, Naihua Xiu
\IEEEcompsocitemizethanks{\IEEEcompsocthanksitem This work is supported by the National Natural Science Foundation of China (No.12131004, No.11971052).
\IEEEcompsocthanksitem  H. Zhang is with School of Management Science, Qufu Normal University, Rizhao Shandong,
China.  S.L. Zhou and N.H Xiu are with the School of Mathematics and Statistics, Beijing Jiaotong University, China. G.Y. Li is with ITP Lab, Department of EEE, Imperial College London, United Kingdom (e-mail: 18118011@bjtu.edu.cn; slzhou2021@163.com; geoffrey.li@imperial.ac.uk; nhxiu@bjtu.edu.cn;).}}

\markboth{Journal of \LaTeX\ Class Files,~Vol.~ , No.~ ,  ~ }%
{Shell \MakeLowercase{{\em et al.}}: 0/1 Deep Neural Networks via Block Coordinate Descent}
%



\maketitle

\begin{abstract}
The step function is one of the simplest and most natural activation functions for deep neural networks (DNNs). As it counts 1 for positive variables and 0 for others, its intrinsic characteristics (e.g., discontinuity and no viable information of subgradients) impede its development for several decades. Even if there is an impressive body of work on designing DNNs with continuous activation functions that can be deemed as surrogates of  the step function, it is still in the possession of some advantageous properties, such as robustness to outliers. Hence,  in this paper, we aim to train DNNs with the step function used as an activation function (dubbed as 0/1 DNNs). We first reformulate 0/1 DNNs as an unconstrained optimization problem and then solve it by a  block coordinate descend (BCD) method. Moreover, we acquire closed-form solutions for sub-problems of BCD as well as its convergence properties. Furthermore, we also integrate $\ell_{2,0}$-regularization into 0/1 DNN to accelerate the training process and compress the network scale.As a result, the proposed algorithm has a desirable performance on classifying MNIST, FashionMNIST, Cifar10, and Cifar100 datasets.
\end{abstract}

%
\IEEEpeerreviewmaketitle

\section{Introduction}
\IEEEPARstart{D}{NN}s can be traced back to the Mcculloch-Pitts (M-P) Neuron \cite{Mcculloch1943}, where the response of a nerve cell follows an ``all-or-none" law \cite{Britannica2019}, that is, the strength of a nerve cell's response is independent of the strength of the stimulus. If a stimulus is above a certain threshold, a nerve will fire. Essentially, there will be either a full response or no response at all for an individual neuron. Such a phenomenon can be characterized by the step function, which returns $1$ if the variable is positive and $0$ otherwise. Therefore, in this paper, we aim to investigate DNNs where the step function is used as an activation function (dubbed as 0/1 activation) on neurons.
\subsection{0/1 DNNs with $\ell_{2,0}$-regularization}
Now we introduce $h$-layer neural networks with $(h-1)$ hidden layers and  the 0/1  activation. Specifically, let $d_i$ be the number of hidden units of the $i$-th hidden layer for $i\in[h-1]$, where $[n]:=\{1,2,\cdots,n\}$ for a positive integer $n$. Let $d_0$ and $d_h$ represent the number of input and output units. Denote $\CW:=\{\W_1,\W_2,\cdots,\W_h\}$ with $\W_i\in \R^{d_{i}\times d_{i-1}}$ being the weight matrix between the $(i-1)$th layer and the $i$th layer. Now we are given a set of input and output data $\{(\x_s,\y_s): s\in[N]\}$, where $(\x_s,\y_s)\in\R^{d_0}\times\R^{d_h}$, $N$ is the batch size of samples and $\y_s$ is a one-hot vector consisting of 0s in all cells with the exception of a single 1 in a cell. The optimization model of 0/1  DNNs for classification problems can be built as follows,
\be\label{OP}
 \arraycolsep=1.4pt\def\arraystretch{1.25}
\ba{cl}
\min\limits_{\CW} &\frac{1}{2N}\sum_{s=1}^N \| \y_s-{\widetilde \y}_s\|^{2} +   g(\CW) \\
{\rm s.t.}& {\widetilde \y}_s=(\W_h(\cdots(\W_2(\W_1\x_s)_{0/1})_{0/1}\cdots)_{0/1})_{\hd},\\
&s\in[N],
\ea
\ee
where $\left\|\cdot\right\|$ represents the Euclidean norm, $g(\CW)$ is a sparsity-induced regularized function. In this paper, we focus on the $\ell_{2,0}$-regularization in $g(\CW)$ for the purpose of getting neural networks with an adaptive width, namely,  
\begin{eqnarray*}
\arraycolsep=1.4pt\def\arraystretch{1.25}
\begin{array}{lll}g(\CW):= \sum_{i=1}^h( \lambda \|\W_{i}\|_{2,0}+\frac{\gamma}{2}\|\W_{i}\|^2 ).\end{array}
\end{eqnarray*}
Here, $\lambda$ and $\gamma$ are positive constants and $\|\W\|_{2,0}$ counts the number of nonzero columns of matrix $\W$. A column is said to be nonzero if it has at least one nonzero entry. For step function $(\a)_{0/1}:\R^n\rightarrow\R^n$ and hardmax  $(\a)_{\hd}:\R^n\rightarrow\R^n$,  their $i$th entries are respectively defined by
\begin{eqnarray*}
\arraycolsep=1.4pt\def\arraystretch{1.25}
\begin{array}{lll}
((\a)_{0/1})_i&=&\left\{\begin{array}{lll}
1,& {\rm if}~ a_i > 0,\\
0,& {\rm otherwise},\\
\end{array}\right.\\
 ((\a)_{\hd})_i&=&\left\{\begin{array}{lll}
1,& {\rm if}~a_i=\max_{r\in[n]} a_r,\\
0,& {\rm otherwise}.
\end{array}\right.\end{array}
\end{eqnarray*}
In the sequel,  the step function is also called the 0/1 function. 
\subsection{Prior Arts} \label{Section-relatedlectures}
To the best of our knowledge, there is no study dedicated exclusively to the optimization analysis of 0/1 DNNs with an $\ell_{2,0}$-regularization. Currently, an impressive body of work has developed various surrogates of the step function as  activation functions. Based on them, a number of algorithms have been cast to train different structured neural networks. In this section, we conduct a brief review of  popular activation functions and relevant algorithms.

\begin{figure*}[!th]
\centering
\includegraphics[width=.85\textwidth]{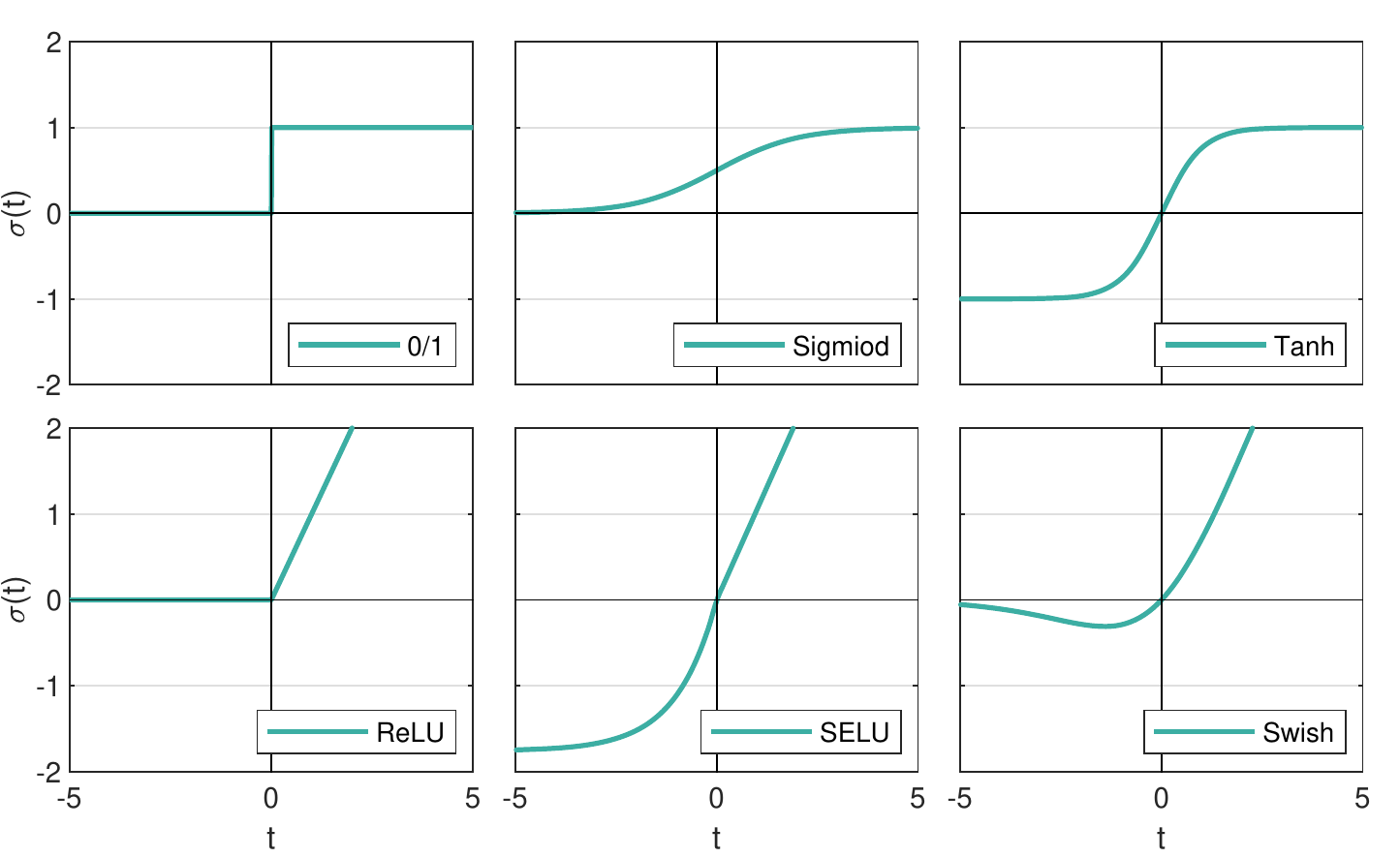}
\caption{Six activation functions.}\vspace{-5mm}
\label{figureactivation}
\end{figure*}

{\bf a) Activation functions} in DNNs play a key role in neural networks so it becomes fundamental to understand their advantages and disadvantages to achieve better performance. One of the simplest and most natural activation functions for DNNs is the step function, which is also regarded as the ideal one in \cite{zhouzhihua}. However, this function is discontinuous and its subgradients vanish almost everywhere. As a consequence, the traditional gradient-driven backpropagation algorithms \cite{Rumelhart1986} cannot be used to train the multilayer networks with the 0/1 activation. Therefore, a plenty of
continuous surrogates have been introduced to approximate it.  We show five popular ones in addition to the step function in Fig. \ref{figureactivation}. One of the widely used surrogates is the Sigmoid function \cite{Han1995}, which was the default activation function used on neural networks for a long time. Another activation function is hyperbolic tangent (Tanh) with a similar shape to Sigmoid.  However, both Sigmoid and Tanh saturate across most of their domain \cite{Goodfellow2016}.
 Since introduced in DNNs \cite{Nair2012}, the Rectifier Linear Unit (ReLU) has quickly gained its popularity due to its desirable features, such as being piecewise linear and avoiding easy saturation. We note that ReLU DNNs are usually learned by the gradient-based algorithms, the family of backpropagation algorithms. However, similar to 0/1 DNNs, ReLU DNNs also involve the problem of ``vanishing gradients" that may prevent backpropagation algorithms from training desired networks. Then SELU (scaled exponential linear unit \cite{Klambauer2017}) and Swish \cite{Ramachandran2017} have been designed to enable very deep neural networks as the issue of vanishing gradients can be  eliminated. Some other popular activation functions include the leaky ReLU \cite{Maas2013},  parametric ReLU \cite{He2015},  ELUs \cite{Clevert2015}, and those in \cite{Goodfellow2016}.

{  A separate line of research aims at exploiting the 1-bit
activation functions, such as the sign and binarization  functions,  resulting in various binary neural networks (BNNs). 
These BNNs also use 1-bit weights. Hence,  they are capable of compressing networks for extreme computational and storage efficiency. However, the discontinuity of 1-bit activation functions makes it enormous difficult to train the BNNs. To overcome this drawback, the popular approach is to approximate these functions or their (sub)gradient. For example,   the authors in \cite{Hinton2012} first proposed a straight-through empirical (STE) gradient estimator approach to train BNNs. The main idea of this method is to approximate the sub-gradient of the sign function by the sub-gradient of the hard tanh function, which enables gradient-driven algorithms to train the networks.  Similar ideas were then extensively adopted by \cite{zhong2012sensitivity,courbariaux2015,Tang2017,Alizadeh2018,Xu-Cheung2019,Gu-Zhang2019,Martinez-Yang2020, Martinez2020, Bulat2021,Ye-wang2021,Zhang-wang2021,jiang-wang2022,Li-Wang2022,Wang-He2022}. For example, methods like compact convolutional neural networks (CCNN \cite{Xu-Cheung2019}) use a derivative estimator to approximate the (sub)gradient of the binarization function, and the Bi-Real-Net \cite{Martinez-Yang2020} employs a polynomial step function to approximate the sign function. We refer to a nice survey for more details \cite{Yuan2021}. Once an algorithm obtains trained parameters based on the approximation, these parameters are then reintroduced into BNNs using 1-bit activations for prediction.}

{We point out that some 1-bit activation functions and the 0/1 function share similar structures, such as being discontinuous at the origin and having zero gradients everywhere except for the origin. Hence the aforementioned approximation methods can be employed to train 0/1 DNNs.   However, some 1-bit activation functions like the sign function have different binary values in comparison with the 0/1 function,  and the sign function is upper semi-continuous while the 0/1 function is lower semi-continuous. Most importantly, differing from the approximation methods, the aim of this paper is to develop a method directly processing the 0/1 function.}

 {\bf b) Gradient descent-based algorithms} have been extensively developed to train neural networks with different continuous activation functions. Popular candidates include  AdaGrad \cite{Duchi2011}, RMSProp \cite{Tieleman2012}, Adadelta \cite{Zeiler2012}, Adam \cite{Kingma2014}, AdaMax \cite{Kingma2014}, Nadam \cite{Dozat2016}, AMSGrad \cite{Reddi2018}, AdamW \cite{Loshchilov2019}, QHAdam \cite{Ma2019},  and AggMo \cite{Lucas2019}. More information  can be found in \cite{Ruder2016} and the references therein. We shall highlight that these algorithms belong to the family of backpropagation algorithms that need to calculate the (sub)gradients of involved functions through a chain rule. As a consequence, the problem of vanishing gradients restricts them from training some DNNs, such as 0/1 DNNs and BNNs. Therefore,  a natural question is if there is an algorithm for 0/1 DNNs without relying on the (sub)gradients heavily like the one proposed in \cite{Taylor2016} so as to get rid of the problem of vanishing gradients.

\subsection{Motivation and contributions}
{
The motivation behind conducting research on 0/1 DNNs is threefold.  Firstly,  similar to BNNs, 0/1 DNNs exploit 1-bit activation functions (i.e., 0/1 functions) and thus enable  network compression for extreme computational and storage efficiency.  Secondly,  the step function offers several advantageous properties. For instance, it can enhance the stability of noises from the training data in DNNs \cite{Courbariaux2015, Courbariaux2016, Yuan2021}. Lastly, despite its drawbacks (such as discontinuity and vanishing gradients) hampering its implementation for backpropagation algorithms for a long time, recent advances have emerged to directly address the 0/1 function, which has led to success in processing various applications, such as the binary classifications \cite{Wang2021, Zhou2021} and one-bit compressive sensing \cite{Zhou12022}. }

{
In addition to employing the step function for both feedforward and backpropagation, we harness the $\ell_{2,0}$-regularization to train the network. It allows us to reduce redundant neurons to get suitable networks to accelerate the computation and mitigate over-fitting issues \cite{lin2019, Yang2019, Dinh2020, Zhang2021, Hoefler2021}. Therefore, inspired by the advantages in utilizing the step function and $\ell_{2,0}$-regularization, our aim is to explore problem \eqref{OP} and make three main contributions.}
 
\begin{itemize}[leftmargin=20pt]
\item[C1)] This is the first paper that investigates  0/1  DNNs with an $\ell_{2,0}$-regularization. To tackle model \eqref{OP}, we first reformulate it as an unconstrained optimization and then develop a block coordinate descend (BCD) method  to solve it from the outer to the inner layer. As the involved sub-problems in BCD either admit closed-form solutions or are easy to be addressed, the proposed method does not belong to the family of backpropagation algorithms. Therefore, there is no problem with the vanishing gradients stemmed from the 0/1 activation. Most importantly, the fulfilment of BCD provides a viable method for 0/1 DNNs, one of the toughest models in deep learning.

\item[C2)] It is noted that problem \eqref{OP} involves three hard functions: the step function, hardmax, and $\ell_{2,0}$-regularization. Hence, it is non-trivial to establish the convergence property for BCD. Nevertheless, we show that any accumulating point of the sequence generated by BCD is a stationary point of the unconstrained optimization problem, see Theorem \ref{the-convergence}. It is worth mentioning that the global convergence of BCD has been established in \cite{Zhang2017, Lau2018, Zeng2019} under some assumptions which, however, are violated by all three hard functions in  \eqref{OP}.

\item[C3)] We conduct some primary numerical experiments for BCD on classifying four datasets. The promising results have demonstrated that 0/1 DNNs not only deliver desirable classification accuracy but also are more robust to adversarial perturbations \cite{Zheng2018} than DNNs with other popular activation functions.
\end{itemize}

\subsection{Organization and notation}
We organize the paper as follows. In the next section, we reformulate problem \eqref{OP} as an unconstrained optimization and describe four types of induced sub-problems. In Section \ref{Section-Technical}, we provide technical analysis for solving the induced sub-problems. It is shown that three of them admit closed-form solutions and one can be solved efficiently. In Section \ref{Section-algorithm}, BCD is developed. Then its implementation and the convergence analysis are also provided. Numerical experiments and conclusions are given in the last two sections.

We end this section by summarizing the notation to be employed throughout the paper.
We use ``$:=$'' to mean ``define''. For an index set $T\subseteq[m]$, let $|T|$ be cardinality of $T$ and $\overline T :=[m]\setminus T$ be its complementary set. Let $\epsilon\to c^+$ represent $\epsilon>c$ and $\epsilon\to c$. Given two vectors $\a,\b\in\R^m$, we use $\am$  to denote the indices of maximal entries of $\a$, namely,
\be
 \arraycolsep=1.4pt\def\arraystretch{1.25}
\ba{lll}
\am:=\{i\in[m]: a_i = \a_{\max}\},~~~\a_{\max}:= \max_{r\in[m]} a_r.
\ea
\ee
Hence,   $\b_{\phi_{\a}}$ means the $\phi_{\a}$th element of $\b$. {It is worth mentioning that $\a$ may have multiple maximal entries and thus $\b_{\phi_{\a}}$ is a vector. However, it is a scalar if $\a$ admits a single maximal value, e.g. one-hot vectors $\y_s$ that we are interested in this paper.} For matrix $\A$, we write $\A_{i:}$, $\A_{:j}$, and $\A_{ij}$ as the $i$th row,  the $j$th column,  and the $(i,j)$th entry of $\A$.
Let $\|\cdot\|$ denote the Frobenius norm for matrices and the Euclidean norm for vectors, and $\|\cdot\|_2$ denote the Spectral norm.  The identity matrix is denoted by  $\I$ and  its  $i$th column is denoted by $\e_i$. The inner product of  two matrices $\A$ and $ \B$ is defined by $\langle \A, \B \rangle:=\sum_{ij} \A_{ij}\B_{ij}$. Finally, for a set of matrices $\{\W_1,\W_2,\cdots,\W_h\}$, denote
\be
\ba{lll}
\CW_{\leq i}:=\{\W_1,\cdots,\W_{i}\},~
 \CW_{\geq i}:=\{\W_{i},\cdots,\W_h\}.
\ea
\ee
\section{Model Reformulation}
We first reformulate original model \eqref{OP} into an unconstrained optimization problem and then briefly describe four types of induced sub-problems.
\subsection{Model reformulation}
We aim to divide multi-composite problem \eqref{OP} into a series of easy-to-analyze sub-problems. To proceed with that, we introduce some intermediate variables as follows. Let ${\U_{i}}=\W_i\V_{i-1}, i\in[h]$ and $\V_i:=(\U_{i})_{0/1}, i\in[h-1]$. Similar to the definition of $\CW$, we denote $\CU:=\{\U_1,\U_2,\cdots,\U_h\}$ and $\CV:=\{\V_1,\V_2,\cdots,\V_{h-1}\}$. Moreover, denote $\V_0:=(\x_1,\x_2,\cdots,\x_N)$ and $\Y:=(\y_1,\y_2,\cdots,\y_N)$ for easier reference. Then problem \eqref{OP} can be rewritten as
\be\label{OP010}
 \arraycolsep=1.4pt\def\arraystretch{1.25}
\ba{cll}
\min\limits_{\CW, \CU, \CV} &&\frac{1}{2N} \|\Y- (\U_h)_{\hd}\|^{2} + g(\CW)\\
{\rm s.t.} &&\U_i=\W_i\V_{i-1},~i\in[h],\\
&&\V_{i}=(\U_i)_{0/1},~~~i\in[h-1],
\ea
\ee
where $(\U_h)_{\hd}{:=}[((\U_h)_{:1})_\hd, \cdots , ((\U_h)_{:N})_\hd]$.

One can discern that it is considerably hard to solve problem \eqref{OP010} (or problem \eqref{OP}). Therefore,  we relax the constraints and shift them to the objective function, resulting in an unconstrained optimization problem,
\be\label{penalf1}
 \arraycolsep=0.0pt\def\arraystretch{1.75}
\begin{array}{lcl}
\min\limits_{\CW,\CU,\CV} F(\CW,\CU,\CV)&:=&  \frac{1}{2N} \|\Y- (\U_h)_{\hd}\|^{2}+ g(\CW)\\
&+&\frac{\tau}{2} \sum_{i=1}^h \|\U_i-\W_{i}\V_{i-1}\|^2 \\
&+& \frac{\pi }{2}\sum_{i=1}^{h-1}\|\V_i-(\U_{i})_{0/1}\|^2,
\end{array}
\ee
where $\tau$ and $\pi$ are positive penalty parameters.

\subsection{Induced sub-problems}\label{sec:induced}
To solving problem \eqref{penalf1}, we will adopt the BCD algorithm, which solves one block while fixing the others. This will induce the following four types of sub-problems.\\
a) The outermost layer sub-problem with respect to $\U_{h}$:
\be\label{U-outproblem}
\arraycolsep=1.0pt\def\arraystretch{1.25}
\ba{l}
\min\limits_{\U_h} \frac{1}{2N}\|\Y- (\U_h)_{\hd}\|^{2}+\frac{\tau}{2} \|\U_h-\W_h\V_{h-1}\|^2.
\ea\ee
b) Inner layer sub-problems with respect to  $\U_i,i\in[h-1]$:
\be\label{U-inproblem}
\ba{cl}
\min\limits_{\U_i}\frac{\tau}{2} \|\U_i-\W_{i}\V_{i-1}\|^2+ \frac{\pi }{2} \|\V_i-(\U_{i})_{0/1}\|^2.
\ea\ee
c) Sub-problems with respect to  $\W_i, i\in[h]$:
\be\label{W-sub-problem}
\ba{cl}
\min\limits_{\W_i} \frac{\tau}{2}\|\U_i-\W_i\V_{i-1}\|^2+\frac{\gamma}{2} \|\W_i\|^2+\lambda \|\W_i\|_{2,0}.
\ea\ee
d) Sub-problems with respect to  $\V_i, i\in[h-1]$:
\be\label{V-sub-problem}
 \arraycolsep=0.4pt\def\arraystretch{1.25}
\ba{crl}
\min\limits_{\V_i}\frac{\tau}{2}\|\U_{i+1}-\W_{i+1}\V_{i}\|^2+ \frac{\pi }{2} \|\V_i-(\U_{i})_{0/1}\|^2.
\ea\ee

\section{Solving Sub-problems} \label{Section-Technical}
In this section,  we aim to solve problems \eqref{U-outproblem}, \eqref{U-inproblem} ,  \eqref{W-sub-problem}, and \eqref{V-sub-problem}. We note that the first two problems  have separable objective functions and thus can be reduced to simpler problems with closed-form solutions, while problem \eqref{W-sub-problem} is slightly hard to tackle but can be solved efficiently.
\subsection{Solutions to sub-problem (\ref{U-outproblem})}
Since  $(\U_h)_{\hd} =[((\U_h)_{:1})_\hd, \cdots , ((\U_h)_{:N})_\hd]$, sub-problem \eqref{U-outproblem} can be solved column-wisely. In addition, as $\Y:=[\y_1,\y_2,\cdots,\y_N]$ and $\y_s,s\in[N]$ are one-hot vectors, to address sub-problem \eqref{U-outproblem},  we can focus on the following problem in a vector form,
\begin{eqnarray}\label{u-outproblem}
 \arraycolsep=1.0pt\def\arraystretch{1.25}
\begin{array}{l}
\u^* \in {\rm arg}\min\limits_{\u\in\R^m} \psi(\u):=\|\y- (\u)_\hd\|^{2}+ \mu \|\u-\b\|^2,
\end{array} \end{eqnarray}
where $\y\in\{\e_1,\e_2,\cdots,\e_m\}\subset\R^m $, $\b\in\R^m$,  and $\mu>0$ are given parameters.  The following lemma states that the above problem admits a closed-form solution.
\begin{lemma}\label{u-outlemma} For any given $\y\in\{\e_1,\e_2,\cdots,\e_m\}$,  solution $\u^*$ to (\ref{u-outproblem}) takes the form of
\begin{eqnarray}\label{soultion-Uh}
 \arraycolsep=1.0pt\def\arraystretch{1.25}
\u^*=  \left\{\begin{array}{lll}
\b, & \mu \Delta^2 > \|\y-(\b)_\hd \|^2,\\
\b~ \text{or}~\b+\epsilon \e_{\phi_\y}, ~~& \mu \Delta^2  = \|\y-(\b)_\hd \|^2,\\
 \b+\epsilon \e_{\phi_\y}, & \mu \Delta^2  < \|\y-(\b)_\hd \|^2,
\end{array}\right.
\end{eqnarray}
where  $\Delta:=\b_{\max}-\b_{\phi_\y}$ and $ \epsilon\to\Delta^+$.
\end{lemma}
 In the numerical experiments, we use $\epsilon=\Delta+10^{-10}$ to approximate $\epsilon\to\Delta^+$. The above result enables us  to show closed-form solution $ \U^*_{h}$ to \eqref{U-outproblem} as follows. For each $s\in[N]$, each column of $ \U^*_{h}$ can be solved by \eqref{soultion-Uh} with setting
\be\ba{l}\label{Uh-problem-*}
(\U^*_{h})_{:s}=\u^*, \y = \y_s,  \b=(\W_h\V_{h-1})_{:s},\mu=\tau N.
\ea\ee
\subsection{Solutions to sub-problem  (\ref{U-inproblem})}
Since the objective function in sub-problem \eqref{U-inproblem} is separable, it can be addressed element-wisely. Therefore, we aim to solve the following one-dimensional problem,
\be\ba{cl}\label{u-inproblem}
u^*\in {\rm arg}\min\limits_{u\in \R}~\varphi (u):= (a- (u)_{0/1})^2+ \rho  (u-b)^2,
\ea\ee
where $a$, $b$, and $\rho >0$ are given scalars. Our next result shows the above problem admits a closed-form solution.
\begin{lemma} \label{u_insolution}Denote $t:=2a-1 $ and $\varepsilon\to0^+$.  Then the solution to  (\ref{u-inproblem}) is solved as follows.\\
i) If $b>0$, then
\begin{eqnarray}
\label{a0-Prox-+}
u^*=
\begin{cases}
b,&t>-\rho b^2,\\
b~\text{or}~0,&t=-\rho b^2,\\
0, &t<-\rho b^2.
\end{cases}
\end{eqnarray}
ii) If $b\leq 0$, then
\begin{eqnarray}
\label{a0-Prox--}
u^*=
\begin{cases}
\varepsilon,&t> \rho b^2,\\
\varepsilon~\text{or}~b,&t= \rho b^2,\\
b, &t<\rho b^2.
\end{cases}
\end{eqnarray}

\end{lemma}

\begin{figure}[!th]
\centering
\includegraphics[width=0.24\textwidth]{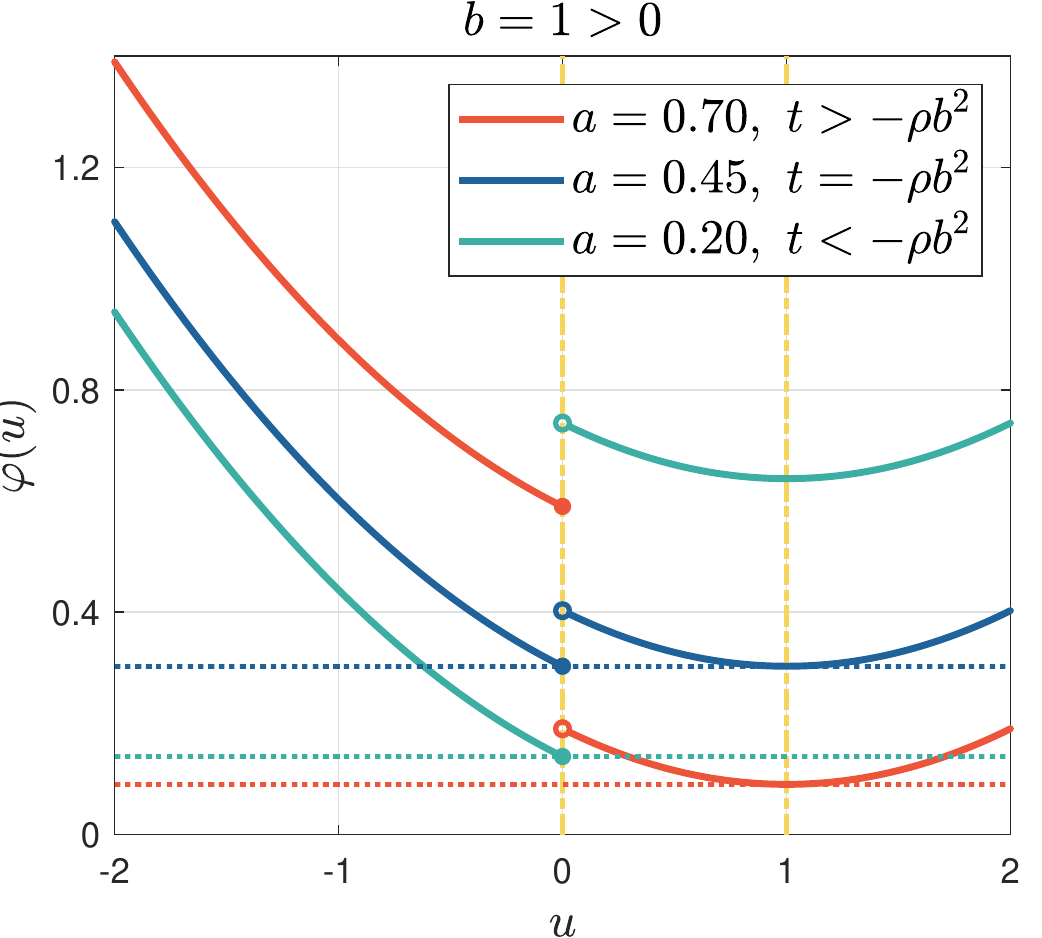}
\includegraphics[width=0.24\textwidth]{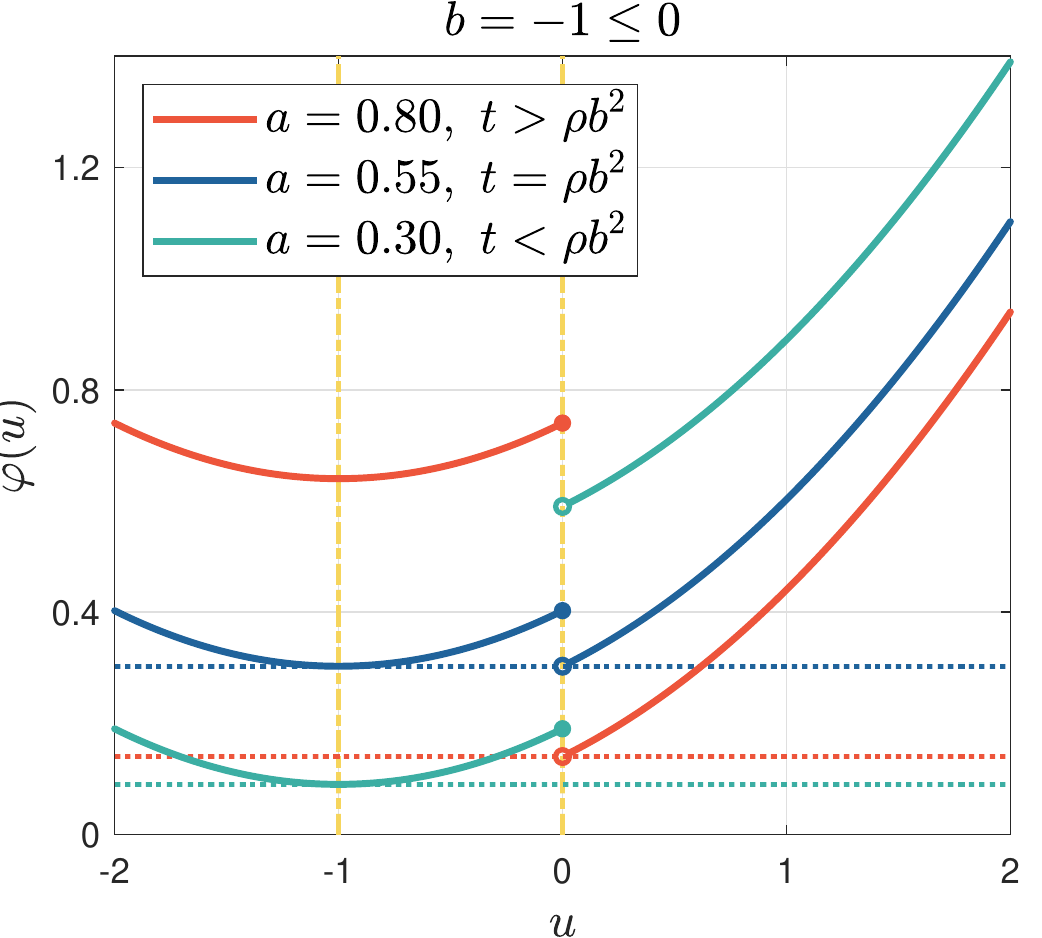}
  \caption{Illustrations for \eqref{a0-Prox-+} and \eqref{a0-Prox--} with $ \rho=0.1, \varepsilon=10^{-10}$.}\label{01-solutions}
\end{figure}

\begin{remark}The solutions to (\ref{u-inproblem}) are illustrated by Fig. \ref{01-solutions}. It is worth mentioning that when $b\leq0$, condition $t>\rho b^2$   means $a>1/2$ and $\sqrt{t/\rho} \geq -b\geq0$, resulting in $\lim_{\epsilon \to0^+}\varphi(\varepsilon) \neq \varphi(0)$. However,  $\varepsilon \to0^+$ is hard to handle numerically. In our numerical experiments, we set $\varepsilon=\min\{\sqrt{t/\rho}+b,10^{-10}\}.$
\end{remark}
The above lemma allows us to derive solution $(\U_i^*)_{jr}=u^*$ to sub-problem \eqref{U-inproblem}  by setting
\be \label{U-inproblem-*}
\ba{lllllll}
a&=&{(\V_i)}_{jr},~ &b&=&(\W_{i}\V_{i-1})_{jr},\\
t&=&2(\V_i)_{jr}-1,~ &\rho&=&\pi/\tau.
\ea\ee
%

\subsection{Solving  sub-problem (\ref{W-sub-problem}) approximately}

Instead of addressing sub-problem \eqref{W-sub-problem} directly, we focus on the following optimization,
\be\label{W-sub-problem31}
\ba{cl}
\W^*\in{\rm argmin}_{\W}~  \Psi(\W)  +\lambda \|\W\|_{0,2},
\ea\ee
where
\be
\ba{cl}
 \Psi(\W):=\frac{\tau}{2}\|\U-\V\W\|^2+\frac{\gamma}{2} \|\W\|^2.\nonumber
\ea\ee
Here, $\U, \V$, and $\W$ correspond to $\U_i^\top, \V_i^\top$, and $\W_i^\top$ in \eqref{W-sub-problem}, and  $\|\W\|_{0,2}:=\|\W^\top\|_{2,0}$ stands for the row sparsity.
%

To solve sub-problem \eqref{W-sub-problem31}, we introduce the concept of a proximal stationary point (P-stationary point), which is associated with the proximal operator in variational analysis \cite{Rockafellar1998}.  Let $G(\cdot):{\mathbb R}^{m\times n}\rightarrow  \R\cup\{+\infty\}$ be a proper lower semi-continuous function. The proximal operator of $G$, associated with  parameter $\beta>0$, is defined by
\begin{eqnarray}\label{def-prox}
\begin{array}{l}
{\rm Prox}_{\beta } G(\H):=\operatorname*{argmin}\limits_{\W\in \R^{m\times n}}~G(\W)+ \frac{1}{2\beta}\|\W-\H\|^2.
\end{array}
\end{eqnarray}
Hu and Beck independently presented the proximal operator of $\|\H\|_{0,2}$ in \cite[Proposition 18]{Hu2017} and \cite[Theorem 3.2]{Beck2019}. Any point  $\W\in{\rm Prox}_{\beta\lambda\|\cdot\|_{0,2}}(\H)$ takes the form of
\begin{eqnarray}\label{Proxl0}
\forall s\in[m]:~~  \W_{s:}=\left\{
 \arraycolsep=1.0pt\def\arraystretch{1.25}
\begin{array}{lll}
0, &{\rm if}&\|\H_{s:}\|<\sqrt{2\beta\lambda},\\
0~{\rm or}~\H_{s:}, &{\rm if}&\|\H_{s:}\|=\sqrt{2\beta\lambda},\\
\H_{s:}, &{\rm if}&\|\H_{s:}\|>\sqrt{2\beta\lambda}.
\end{array} \right.
\end{eqnarray}
Based on the proximal operator, the P-stationary point of  \eqref{W-sub-problem31} can be defined as follows.
\begin{definition}\label{def-S-prox2}
A matrix $\W^*\in\R^{m\times n}$ is a P-stationary point of  problem (\ref{W-sub-problem31}) if there is a constant $\beta>0$ such that
\be\label{P-CW}
    \W^* \in {\rm Prox}_{\beta \lambda\|\cdot\|_{0,2}}(\W^*-\beta \nabla \Psi(\W^*)).\ee
\end{definition}
It is easy to show that $\W^*$ is a P-stationary point if and only if it satisfies that for any $s\in[m]$,
 \be \label{P-equivalent}
  \arraycolsep=1.4pt\def\arraystretch{1.25}
\left\{ \ba{lll}
\|(\nabla \Psi(\W^*))_{s:}\|=0,~ \|\W^*_{s:}\|\geq
\sqrt{2\beta\lambda}, & \text{if} ~ \|\W^*_{s:}\|\neq 0, \\
\|(\nabla \Psi(\W^*))_{s:}\|\leq \sqrt{2\lambda/\beta}, & \text{if} ~ \|\W^*_{s:}\|= 0.
 \ea \right.
 \ee
In the subsequent part, we first show that a P-stationary point is a unique locally optimal solution  to problem \eqref{W-sub-problem31} and satisfies a quadratic growth property.
 \begin{theorem} \label{P-first-order-local}
Let $\W^*$ be a P-stationary point for some $\beta>0$, then it is a unique locally optimal solution to problem (\ref{W-sub-problem31}) and satisfies the following quadratic growth property,
 \be \label{quadratic-growth-property}
  \arraycolsep=1.4pt\def\arraystretch{1.25}
\ba{lll}
&&\Psi(\W) + \lambda\|\W\|_{0,2} -( \Psi(\W^*) + \lambda\|\W^*\|_{0,2})\\
&  \geq & \frac{\gamma}{2} \|\W -\W^*\|^2,~~\forall ~\W\in{\mathbb N}(\W^*, \sqrt{\beta\lambda/(2m^2)})
 \ea
 \ee
where $ {\mathbb N}(\W^*, \delta):= \{\W: \|\W -\W^*\|<\delta\}$.
\end{theorem}
In addition to the relation to the locally optimal solution,
 a P-stationary point also has a close relationship with the globally optimal solution to   \eqref{W-sub-problem31}, stated as follows.
\begin{theorem} \label{P-first-order}
If $\W^*$ is a globally optimal solution to problem (\ref{W-sub-problem31}), then it is a P-stationary point for any $\beta\in(0,1/(\tau \|\V\|_2^2+\gamma))$. On the contrary, a P-stationary point with $\beta\geq 1/\gamma$ is also a globally optimal solution.
\end{theorem}
The above theorem shows that instead of finding an optimal solution to \eqref{W-sub-problem31}, it is meaningful to find a P-stationary point as the later can be achieved by the proximal gradient method (PGM) \cite{Beck2019} presented in Algorithm \ref{ses}.
 \begin{algorithm}[H] %
	\caption{PGM($ \gamma,  \lambda, \U, \V, \W^0$)}\label{ses}
	\begin{algorithmic}[1]
    \STATE \textbf{Initialize}   $L>0$ and  $\beta>0$.
	\FOR{$\ell=0,1,\cdots,L$}	
	 \STATE  Update  $\W^{\ell+1}$ by
	  \be\label{P-CW-l}
    \W^{\ell+1} \in {\rm Prox}_{\beta \lambda\|\cdot\|_{0,2}}(\W^{\ell}-\beta \nabla \Psi(\W^{\ell})).\ee
\ENDFOR
\RETURN $\W^{\ell+1}$
	\end{algorithmic}
\end{algorithm}
Algorithm \ref{ses}  is associated with $\gamma, \U$, and $ \V$  since it aims to deal with \eqref{W-sub-problem31}. Here,  $\W^0$ is an initial point that can be set by the previous point during the training.  The following theorem shows that the whole sequence generated by Algorithm \ref{ses}  converges to a  P-stationary point of  \eqref{W-sub-problem31}.
\begin{theorem} \label{PGM-converge}
Let $\{\W^\ell\}$ be the sequence generated by Algorithm \ref{ses} with $\beta\in(0,1/(\tau \|\V\|_2^2+\gamma))$, then the whole sequence converges to a P-stationary point of problem (\ref{W-sub-problem31}).
\end{theorem}

\subsection{Solving  sub-problem (\ref{V-sub-problem}) }
{Problem (\ref{V-sub-problem}) can be solved by finding  solution $\V_i^*$ to the following linear equation,
\be\label{V-sub-problem-sol}
(\tau\W_{i+1}^\top\W_{i+1}+\pi \I)\V_i^*=\tau\W_{i+1}^\top\U_{i+1} +\pi(\U_{i})_{0/1}.
\ee
One way is to compute the inverse of $(\tau\W_{i+1}^\top\W_{i+1}+\pi \I)$ with computational complexity $O(d_{i-1}^3)$, which is impractical when $d_{i-1}$ is large.  Alternatively, we can adopt the conjugate gradient method, which requires computational complexity $O(td_{i-1}^2)$, where $t$ is the iteration number and usually small. In our numerical experiments, we adopt this method to solve sub-problem (\ref{V-sub-problem}).}
%
\section{Block Coordinate Descent} \label{Section-algorithm}
In this section, we adopt the BCD  algorithm  to solve problem \eqref{penalf1} and then establish its convergence property.

\subsection{Algorithmic design}

Suppose we have computed $(\CU^{k}, \CW^{k}, \CV^k)$ at step $k$. Then next point   $(\CU^{k+1}, \CW^{k+1},\CV^{k+1})$ is updated as \eqref{Int2}.

\begin{figure}[!th]
\centering
\includegraphics[width=1\linewidth]{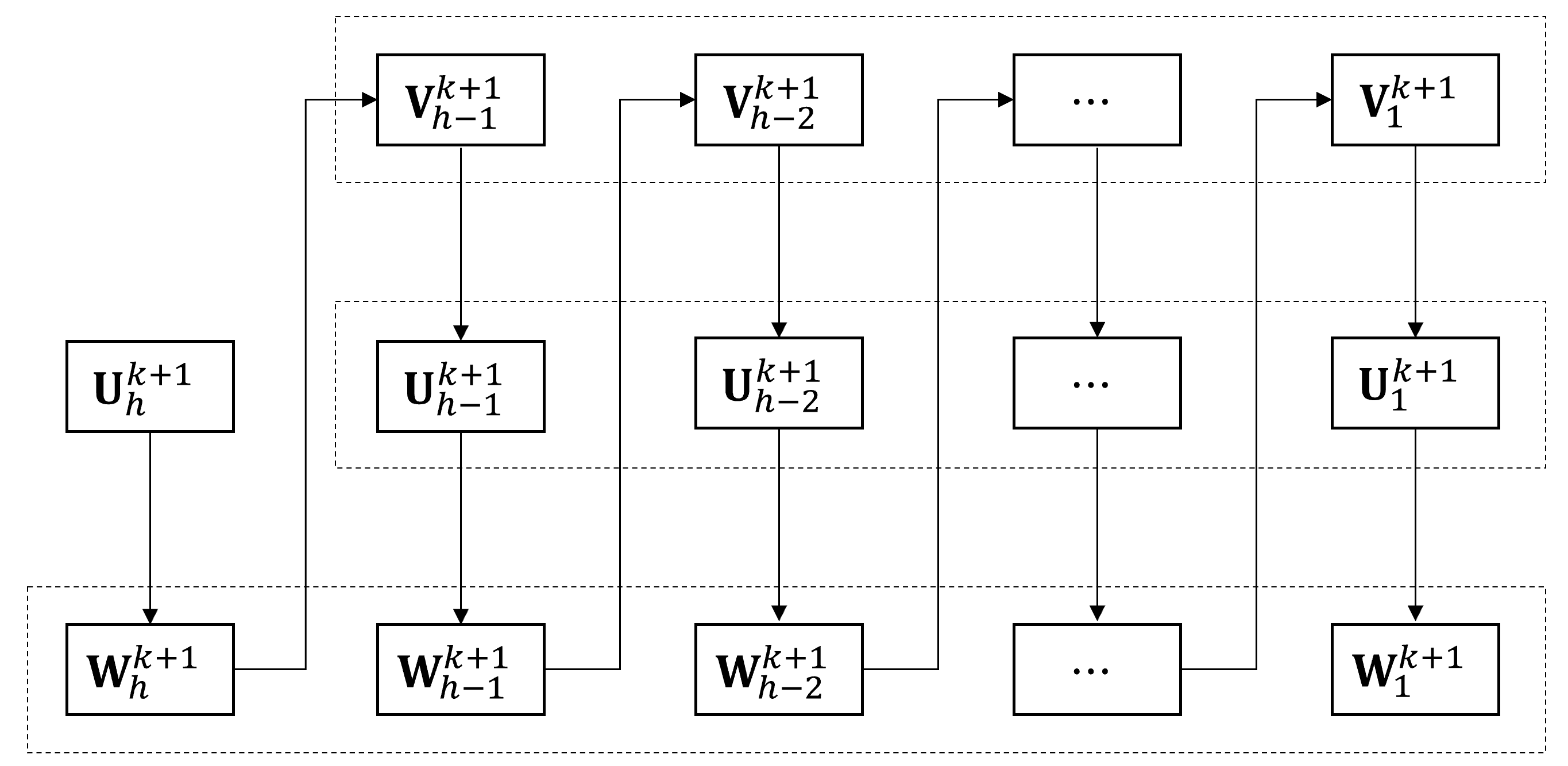}
  \caption{{Update order of parameters.}}\label{scheldual}
\end{figure}

\begin{table*}[!htb]
\centering
\begin{tabular}{lcccr}
\parbox{1\textwidth}{ \begin{eqnarray}
\arraycolsep=1.4pt\def\arraystretch{1.25}
\label{Int2}
&&\left\{\ba{lcll}
\U_h^{k+1}&\in&\operatorname{arg} \min\limits_{\U_h}& F(\CW^{k}, \{\CU_{\leq h-1}^{k},\U_h\}, \CV^{k}),\\
\W_h^{k+1}&\in&\operatorname{arg} \min\limits_{\W_h}& F(\{\CW_{\leq h-1}^{k},\W_h\}, \{\CU_{\leq h-1}^{k},\U_h^{k+1}\}, \CV^k),\\
\V_{h-1}^{k+1}&=&\operatorname{arg} \min\limits_{\V_{h-1}}& F(\{\CW_{\leq h-1}^{k},\W_h^{k+1}\}, \{\CU_{\leq h-1}^{k},\U_h^{k+1}\}, \{\CV_{\leq h-2}^{k},\V_{h-1}\}),\\
\U_{h-1}^{k+1}&\in&\operatorname{arg} \min\limits_{\U_{h-1}}& F(\{\CW_{\leq h-1}^{k},\W_h^{k+1}\},  \{\CU_{\leq h-2}^{k},\U_{h-1},\U_h^{k+1}\}, \{\CV_{\leq h-2}^{k},\V_{h-1}^{k+1}\}),\\
\W_{h-1}^{k+1}&\in&\operatorname{arg} \min\limits_{\W_{h-1}}& F(\{\CW_{\leq h-2}^{k},\W_{h-1},\W_h^{k+1}\},  \{\CU_{\leq h-2}^{k},\CU_{\geq h-1}^{k+1}\}, \{\CV_{\leq h-2}^{k},\V_{h-1}^{k+1}\}),\\
&\vdots&\\
\V_{1}^{k+1}&=&\operatorname{arg} \min\limits_{\V_1}& F(\{\W_{1}^{k},\CW_{\geq 2}^{k+1}\},  \{\U_{1}^{k},\CU_{\geq 2}^{k+1}\}, \{\V_1, \CV_{ \geq 2 }^{k+1}\}),\\
\U_{1}^{k+1}&\in&\operatorname{arg} \min\limits_{\U_1}& F(\{\W_{1}^{k},\CW_{\geq 2}^{k+1}\}, \{\U_1,\CU_{\geq 2}^{k+1}\}, \CV^{k+1}),\\
\W_{1}^{k+1}&\in&\operatorname{arg} \min\limits_{\W_1}& F(\{\W_{1},\CW_{\geq 2}^{k+1}\}, \CU^{k+1}, \CV^{k+1}).
\ea\right.\\
\label{G-In3}
&& \left\{\ba{lclll}
\U_{h}^*&\in&\operatorname{arg} \min\limits_{\U_h}  &\frac{1}{2N} \|\Y- (\U_h)_{\hd}\|^{2}+ \frac{\tau}{2} \|\U_h-\W_{h}^*\V_{h-1}^*\|^2,\\
\W_i^*&\in& {\rm Prox}_{\beta \lambda\|\cdot\|_{2,0}}&\left(\W_i^*-\beta \nabla_\W \Phi ( \W_{i}^*; \U_{i}^*,\V_{i-1}^*)  \right), & \forall i\in[h],\\
\V_{i}^*&=&\operatorname{arg} \min\limits_{\V_{i}}& \frac{\tau}{2}\|\U_{i+1}^*-\W_{i+1}^*\V_{i}\|^2+ \frac{\pi}{2}\|\V_{i}-(\U_{i}^*)_{0/1}\|^2, &\forall i\in[h-1],\\
\U_i^*&\in&\operatorname{arg} \min\limits_{\U_i}& \frac{\tau}{2}\|\U_i-\W_i^*\V_{i-1}^*\|^2+  \frac{\pi}{2}\|\V_i^*-(\U_i)_{0/1}\|^2, &\forall i\in[h-1].\\
\ea\right.
\end{eqnarray}} \\\hline
\end{tabular}
\end{table*}
\indent Based on the order of updating each variable in Fig. \ref{scheldual}, the updating starts from the outermost layer to the first layer. By dropping constant terms, the sub-problems in \eqref{Int2} are equivalent to following ones,
\begin{align}
& \label{Int3-Uh}
\arraycolsep=0pt\def\arraystretch{1.25}
\ba{lcll}
\U_{h}^{k+1}~\in\operatorname{arg} \min\limits_{\U_h}  &&\frac{\tau}{2} \|\U_h-\W_{h}^{k}\V_{h-1}^{k}\|^2+\\
&&\frac{1}{2N} \|\Y- (\U_h)_{\hd}\|^{2} ,\ea \\
& \label{Int3-Wi}
\arraycolsep=0pt\def\arraystretch{1.25}
\ba{lcll}
\W_{i}^{k+1}\in\operatorname{arg} \min\limits_{\W_i}  &&\frac{\tau}{2} \|\U_{i}^{k+1}-\W_{i}\V_{i-1}^k\|^2+\\
&&\frac{\gamma }{2}\|\W_{i}\|^2+ \lambda \|\W_{i}\|_{2,0},~ i\in[h], \ea \\
& \label{Int3-Vi}
\arraycolsep=0pt\def\arraystretch{1.25}
\ba{lcll}
\V_{i}^{k+1}~=\operatorname{arg} \min\limits_{\V_{i}}  &&\frac{\tau}{2}\|\U_{i+1}^{k+1}-\W_{i+1}^{k+1}\V_{i}\|^2+\\
&&\frac{\pi}{2}\|\V_{i}-(\U_{i}^{k})_{0/1}\|^2,~ i\in[h-1],  \ea \\
& \label{Int3-Ui}
\arraycolsep=0pt\def\arraystretch{1.25}
\ba{lcll} \U_{i}^{k+1}~\in\operatorname{arg} \min\limits_{\U_{i}}
&&\frac{\tau}{2}\|\U_{i}-\W_{i}^{k}\V_{i-1}^{k}\|^2+\\
&& \frac{\pi}{2}\|\V_{i}^{k+1}-(\U_{i})_{0/1}\|^2,~ i\in[h-1]. \ea
\end{align}
Although there are $(3h-1)$ sub-problems to be solved in \eqref{Int2}, we only need to solve 4 types of optimization problems essentially, namely,  4 types of the induced sub-problems in the above scheme, also see Section \ref{sec:induced}.

Recalling the relationship between a globally optimal solution and a P-stationary point in Theorem \ref{P-first-order}, instead of solving  \eqref{Int3-Wi}, we aim to find a P-stationary point by solving
\begin{eqnarray}
 \label{Int3-Wi-k} 
\ba{lcll}
 \W_{i}^{k+1} \in   {\rm Prox}_{\beta \lambda\|\cdot\|_{2,0}}\Big( \W_i^{k+1}-\beta \nabla_\W \Phi\Big), \ea
\end{eqnarray}
for $i\in[h]$, where  $\nabla_\W \Phi:=\nabla_\W \Phi ( \W_{i}^{k+1}; \U_{i}^{k+1},\V_{i-1}^k)$ is the gradient of  $\Phi(\cdot)$ defined by
\begin{eqnarray}
 \label{Int3-Wi-obj}\ba{lcll}
 \Phi(W;U,V):=\frac{\tau}{2} \|\U-\W\V\|^2+\frac{\gamma }{2}\|\W\|^2.
 \ea
\end{eqnarray}
We point out that \eqref{Int3-Wi-k} can be solved by Algorithm \ref{ses} due to Theorem \ref{PGM-converge}. Now,   all steps for updating $(\CU^{k}, \CW^{k}, \CV^k)$ are summarized in Algorithm \ref{BCD-0}.
 \begin{algorithm}[!th] 
	\caption{BCD for 0/1 DNNs\label{BCD-0}}
	\begin{algorithmic}[1]
	\STATE \textbf{Require} training data $\{(\x_s,\y_s)\}, s\in[N]$ and network parameters $(h, N, d_1,d_2,\cdots,d_h)$.
	\STATE \textbf{Initialize} $\CU^{0}, \CW^{0}, \CV^0$, $\gamma, \lambda, \tau, \pi, \beta>0$, and maximum number of iteration $K$.
	\FOR{$k=0,1,2,\cdots,K$}
	\STATE Update $\U_h^{k+1}$ by \eqref{Int3-Uh}.
	\STATE Update $\W_h^{k+1}$ by \eqref{Int3-Wi-k}, namely, by Algorithm \ref{ses}.
	 \FOR{ $i=h-1, h-2,\cdots,1$}
		\STATE Update $\V_i^{k+1}$ by \eqref{Int3-Vi}.
		\STATE Update $\U_i^{k+1}$ by \eqref{Int3-Ui}.
		\STATE Update $\W_i^{k+1}$ by \eqref{Int3-Wi-k}, namely, by Algorithm \ref{ses}.
		\ENDFOR
\ENDFOR
\RETURN $\CW^{k+1}$
	\end{algorithmic}
\end{algorithm}

\subsection{Convergence analysis}



To establish the convergence property, we introduce the stationary point of problem \eqref{penalf1}.
\begin{definition}\label{def-Gstationary}
A point  {($\CW^*; \CU^*; \CV^*$)} is a stationary point of  problem (\ref{penalf1}) if there exists a constant $\beta>0$ satisfying (\ref{G-In3}).
\end{definition}
The next theorem shows that the objective function values of the sequence  generated by Algorithm \ref{BCD-0} converge and any its accumulating point is a stationary point.
\begin{theorem}\label{the-convergence}
Let $\{(\CW^k; \CU^k; \CV^k)\}$ be the sequence generated by Algorithm \ref{BCD-0}. Then  the following statements are valid.
\begin{itemize}[leftmargin=15pt]
\item[a)] For any $k\geq 0$,
\be\label{decreasing-pro}
  \arraycolsep=1.4pt\def\arraystretch{1.75}
\ba{lcl}
&& F(\CW^{k+1},\CU^{k+1},\CV^{k+1})- F(\CW^k,\CU^k,\CV^k)\\
&\leq&- \frac{\gamma-1/\beta}{2}\sum_{i=1}^n \|  \W_{i}^{k+1}-\W_{i}^{k}\|^2\\ &&-\frac{\pi}{2}\sum_{i=1}^{h-1} \|\V_{i}^{k+1}-\V_{i}^{k} \|^2.
\ea\ee
\item[b)] For any $\gamma\geq1/\beta$, sequence $ \{ F(\CW^{k},\CU^{k},\CV^{k}) \}$ is non-increasing and converges. Hence,
\begin{eqnarray}\label{gap-0}
  \arraycolsep=1.4pt\def\arraystretch{1.25}
\ba{lcl}
\lim\limits_{k\to\infty}(\CW^{k+1}-\CW^k) = \lim\limits_{k\to\infty}(\CV^{k+1}-\CV^k)=0.
\ea\end{eqnarray}
\item[c)] Let $\gamma{>}1/\beta$. Then any accumulating point  $(\CW^*,\CU^*,\CV^*)$   is a  stationary point of problem (\ref{penalf1}) if every column of $\W_h^*\V_{h-1}^*$ only has one maximal entry.
\end{itemize}
\end{theorem}

\subsection{Implementation of BCD}
In the sequel, we present the details for each sub-problem in Algorithm \ref{BCD-0}  so as to make the algorithm tractable.

\begin{itemize}[leftmargin=10pt]
\item \underline{$\U_h^{k+1}$ sub-problem} can be solved by
\begin{eqnarray}\label{soultion-Uh-k1}
 \arraycolsep=0pt\def\arraystretch{1.25}
 (\U_h^{k+1})_{:s}=  \left\{\begin{array}{lll}
\b^{k}_s, & \tau  (\Delta_s^k)^2 > \delta_s^k,\\
\b^{k}_s~ \text{or}~\b^{k}_s+\epsilon^{k}_s \e_{\phi_{\y_s}}, ~~& \tau (\Delta_s^k)^2  = \delta_s^k,\\
 \b^{k}_s+\epsilon^{k}_s \e_{{\phi_{\y_s}}}, & \tau  (\Delta_s^k)^2  < \delta_s^k,
\end{array}\right.
\end{eqnarray}
for $\forall s\in[N]$ due to \eqref{soultion-Uh} and \eqref{Uh-problem-*}, where
\begin{eqnarray*}
 \arraycolsep=1.4pt\def\arraystretch{1.75}
\begin{array}{l}
\b^{k}_s:=(\W_h^{k}\V_{h-1}^{k})_{:s},~\Delta_s^k:=(\b^{k}_s)_{\max}-(\b^{k}_s)_{\phi_{\y_s}}, \\
\delta_s^k:=\frac{1}{N}\|\y_s-(\b^{k}_s)_\hd \|^2,~ \epsilon^{k}_s\to(\Delta_s^k)^+.
\end{array}
\end{eqnarray*}
\item \underline{$\W_i^{k+1}$ sub-problem} can be solved by Algorithm \ref{ses}, i.e.,
\begin{eqnarray}\label{soultion-Wi-k1}
 \arraycolsep=0pt\def\arraystretch{1.25}
 \begin{array}{l}
 \W_i^{k+1}= (\text{PGM}( \gamma, \lambda, (\U_i^{k+1})^{\top}, (\V_{i-1}^k)^{\top},  (\W^k_{i} )^{\top}))^\top.
\end{array} \end{eqnarray}
for $\forall i\in[h]$. As we mentioned before, we use $(\W^k_{i} )^{\top}$ as the initial point for Algorithm \ref{ses}.
\item \underline{$\V_i^{k+1}$ sub-problem} can be solved by finding solution $\V_i^{k+1}$ to the following equation for  $\forall i\in[h-1]$,
\begin{eqnarray}\label{soultion-Vi-k1}
 \arraycolsep=1.4pt\def\arraystretch{1.25}
 \begin{array}{lll}
&&( \tau(\W_{i+1}^{k+1})^\top \W_{i+1}^{k+1}+\pi \I ) \V_i^{k+1}\\
&=&   \tau(\W_{i+1}^{k+1})^\top\U_{i+1}^{k+1} +\pi(\U_{i}^k)_{0/1}.
 \end{array}\end{eqnarray}
\item \underline{$\U_i^{k+1}$ sub-problem} can be solved as follows for $\forall i\in[h-1]$ and  $\forall (j,r) \in[d_i]\times[N]$. \\
If $b_{jr}^k>0$ then
\begin{eqnarray}
\label{soultion-Ui-k1}
 \arraycolsep=1.0pt\def\arraystretch{1.25}
{[\U_i^{k+1}]_{jr}}=
\left\{
\begin{array}{lll}
b_{jr}^k,& \pi t_{jr}^{k+1}>-\tau (b_{jr}^k)^2,\\
b_{jr}^k~\text{or}~0,~~&  \pi t_{jr}^{k+1} =-\tau (b_{jr}^k)^2,\\
0, &\pi t_{jr}^{k+1} <-\tau (b_{jr}^k)^2.
\end{array} \right.
\end{eqnarray}
If $b_{jr}^k\leq0$ then
\begin{eqnarray}
\label{soultion-Ui-k2}
 \arraycolsep=1.0pt\def\arraystretch{1.25}
{[\U_i^{k+1}]_{jr}}=
\left\{
\begin{array}{lll}
\varepsilon,&  \pi t_{jr}^{k+1} >~~\tau (b_{jr}^k)^2,\\
\varepsilon~\text{or}~b_{jr}^k,~& \pi t_{jr}^{k+1}=~~\tau (b_{jr}^k)^2,\\
b_{jr}^k, &\pi t_{jr}^{k+1}<~~\tau (b_{jr}^k)^2,
\end{array}\right.
\end{eqnarray}
due to \eqref{a0-Prox-+}, \eqref{a0-Prox--}, and \eqref{U-inproblem-*}, where
\begin{eqnarray*}
 \arraycolsep=1.0pt\def\arraystretch{1.25}
\begin{array}{lll}
t_{jr}^{k+1} :=  2(\V_i^{k+1})_{jr}-1,~~
 b_{jr}^k := (\W_{i}^k\V_{i-1}^k)_{jr},~~
 \varepsilon \to 0^+.
 \end{array}
\end{eqnarray*}
\end{itemize}
Overall, the detailed computations for $(\CU^{k}, \CW^{k}, \CV^k)$ are summarized in Algorithm \ref{BCD}.

 \begin{algorithm}[!th] 
	\caption{BCD for 0/1 DNNs\label{BCD}}
	\begin{algorithmic}[1]
	\STATE \textbf{Require} training data $\{(\x_s,\y_s): s\in[N]\}$ and network parameters $(h, N, d_1,d_2,\cdots,d_h)$.
	\STATE \textbf{Initialize} $\CU^{0}, \CW^{0}, \CV^0$, $\gamma, \lambda, \tau, \pi>0$, and maximum number of iteration $K$.
	\FOR{$k=0,1,2,\cdots,K$}
	\STATE Update $\U_h^{k+1}$ by \eqref{soultion-Uh-k1}.
	\STATE Update $\W_h^{k+1}$ by \eqref{soultion-Wi-k1}.
	 \FOR{ $i=h-1, h-2,\cdots,1$}
		\STATE Update $\V_i^{k+1}$ by solving \eqref{soultion-Vi-k1}.
		\STATE Update $\U_i^{k+1}$ by \eqref{soultion-Ui-k1} and \eqref{soultion-Ui-k2}.
		\STATE Update $\W_i^{k+1}$ by \eqref{soultion-Wi-k1}.
		\ENDFOR
\ENDFOR
\RETURN $\CW^{k+1}$
	\end{algorithmic}
\end{algorithm}
 
\subsection{Some extensions}\label{extension-BCD}
We extend BCD to train networks in some other scenarios. 
\begin{itemize}[leftmargin=12pt]
\item[I.] If the popular softmax and cross entropy (CE) are used in the network as the outermost activation functions and the  loss function, we can amend model (\ref{OP010}) as 
\begin{eqnarray}\label{OP010-s}
 \arraycolsep=0pt\def\arraystretch{1.25}
\ba{cll}
\min\limits_{\CW, \CU, \CV} &&\sum_{s=1}^N CE(\y_s,((\W_h\V_{h-1})_{:s})_{\rm sfmax})  + g(\CW)\\
{\rm s.t.} &&\U_i=\W_i\V_{i-1},~i\in[h-1],\\
&&\V_{i}=(\U_i)_{0/1},~~~i\in[h-2],
\ea
\end{eqnarray}
where $CE(\y,\a):=-\sum_{i=1}^{d}y_i\log(a_i)$ is the cross entropy between $\y\in\R^{d}$ and $\a\in\R^{d}$, and  $(\a)_{\rm sfmax}:=(e^{a_1},\cdots,e^{a_{d}})^\top/\sum_{i=1}^d e^{a_i}$ is the softmax. Accordingly,  in Algorithm \ref{BCD} we can remove the subproblem of $\U_h$ and replace two subproblems  of $\W_h$ and $\V_{h-1}$ by
\begin{eqnarray*}
\label{U-814problem}
\arraycolsep=1.0pt\def\arraystretch{1.25}
\ba{r}
\min\limits_{\W_h} \sum_{s=1}^N CE(\y_s,((\W_h\V_{h-1})_{:s})_{\rm sfmax})\\
+\lambda \|\W_{h}\|_{2,0}+\frac{\gamma}{2}\|\W_{h}\|^2,\\
\min\limits_{\V_{h-1}}\sum_{s=1}^N CE(\y_s,((\W_h\V_{h-1})_{:s})_{\rm sfmax})\\
+ \frac{\pi }{2} \|\V_{h-1}-(\U_{h-1})_{0/1}\|^2.
\ea\end{eqnarray*}
Both problems can be solved by some popular optimization algorithms, such as the gradient descent (GD) method.
\item[II.]  BCD can be adjusted to train a  convolutional neural network (CNN). Given a CNN with $h^c$ convolutional layers and $h$ fully connected layers where the 0/1 function is used as the activation functions, let $\V_0^c=\V_0$ be the input and 
\begin{eqnarray*} 
 \arraycolsep=0pt\def\arraystretch{1.25}
\ba{l}
\CW:=(\W_1^c,\cdots,\W_{h^c}^c,\W_1,\cdots,\W_{h}),\\\CU:=(\U_1^c,\cdots,\U_{h^c}^c,\U_1,\cdots,\U_{h}),\\\CV:=(\V_1^c,\cdots,\V_{h^c-1}^c,\V_1,\cdots,\V_{h-1}).
\ea
\end{eqnarray*}
Then to train CNN, we can adjust model (\ref{OP010}) as follows
\begin{eqnarray}\label{OP010-s-c}
 \arraycolsep=0pt\def\arraystretch{1.25}
\ba{cll}
\min\limits_{\CW, \CU, \CV} &&\frac{1}{2N} \|\Y- (\U_h)_{\hd}\|^{2}   + g(\CW)\\
{\rm s.t.} 
&&\U_i^c=\W_i^c\V_{i-1}^c, \V_{i}^c=\sigma(\U_i^c), i\in[h^c], \\
&&\U_1=\W_1\V_{h^c}^c,\U_i=\W_i\V_{i-1},i(\neq1)\in[h],\\
&&\V_{i}=(\U_i)_{0/1},i\in[h-1],\\
\ea
\end{eqnarray}
where $\W_i^c$ corresponds to the kernel in the $i$th convolutional layer, $\U_i^c$ represents the output after the kernel operation on input $\V_i^c$,   and $\sigma(\cdot)$ can be deemed as the pooling operation. Therefore, in BCD two additional subproblems of $\U_i^c$ and $\V_i^c$ are induced as follows:
\begin{eqnarray*}
 \arraycolsep=1pt\def\arraystretch{1.25}
\ba{cll}
\min\limits_{\U_i^c} &&\frac{\tau}{2} \|\U_i^c-\W_i^c\V_{i-1}^c\|+ \frac{\pi}{2}\| \V_i^c-\sigma(\U_i^c)\|^{2},\\
\min\limits_{\V_i^c} &&\frac{\tau}{2} \|\U_{i+1}^c-\W_{i+1}^c\V_i^c\|+ \frac{\pi}{2}\| \V_i^c-\sigma(\U_{i+1}^c)\|^{2},\\
\ea
\end{eqnarray*}
where the first problem may have closed form solutions for some particular pooling functions  $\sigma(\cdot)$ (e.g., max pooling) and the second one can be addressed similarly to  (\ref{V-sub-problem}).  

Moreover, combining model \eqref{OP010-s} and \eqref{OP010-s-c}, BCD enables training CNN with softmax used as the outermost activation functions   and cross entropy used for the loss function.
\end{itemize}
 
\section{Numerical Experiments} \label{Section-Numerical}
In this section, we will conduct some numerical experiments to demonstrate the performance of BCD 
in Algorithm \ref{BCD} using python on Collab provided by Google (CPU version). To show the efficiency of the 0/1 activation, we will compare 0/1 DNNs (solved by BCD) with DNNs with different popular activation functions (solved by Adam \cite{Kingma2014})  for classifying four datasets.
\subsection{Datasets and implementations}
We aim to implement our algorithm to classify four popular datasets: MNIST \cite{LeCun2010}, FashionMNIST \cite{Xiao2017}, { Cifar10 \cite{Krizhevsky-Nair2014}, and Cifar100 \cite{Krizhevsky-Nair2014}}. For each dataset we build the following architecture of 0/1 DNNs.

\begin{itemize}[leftmargin=15pt]
\item[i)] MNIST dataset with 10 classes has 60,000 and 10,000 images used for training and testing. Each piece of data is a 28$\times$28 grey-scale handwritten digit image. For such a dataset, we train our 0/1 DNNs with two hidden layers and one  output layer. {All layers are fully-connected. We denote this network by F3NN.} The number of each hidden layer node is set as 2000. Overall, we have $(h, N, d_1,d_2,d_3)=(3, 60000, 2000, 2000, 10)$.

\item[ii)] FashionMNIST dataset with 10 classes consists of 28 $\times$ 28 grayscale images curled from 70,000 fashion products \cite{Xiao2017}. Again, it is divided into two parts: 60,000 images for training and 10,000 images for testing. The architecture of our 0/1 DNNs is similar to the one for the MNIST dataset. Therefore,  $(h, N, d_1,d_2,d_3)=(3, 60000, 2000, 2000, 10)$.

\item[iii)] { Both Cifar10  and Cifar100 datasets comprise 50,000 images for training and 10,000 images for testing. Each image is a full-color $32\times32$ picture of real objects such as airplanes, cars, and birds. Although the sample size is slightly smaller than that of MNIST and FashionMNIST,  the images have much more features, as $32^2\times3 > 28^2$.

To classify these two datasets, we leverage the architecture of Resnet18, a popular type of CNN.  The descriptions of BCD to train this type of networks were given in Section \ref{extension-BCD}.  Finally, to classify the Cifar10, we use hardmax at the the output layer along with the mean squared error (MSE) loss in \eqref{OP010-s-c}. However, to classify the Cifar100, we use the softmax and CE as shown in model \eqref{OP010-s}. This choice highlights the flexibility of our BCD approach in handling DNNs with diverse activation and loss functions.}
\end{itemize}
\begin{figure*}[!th]
\centering
\includegraphics[width=1\textwidth]{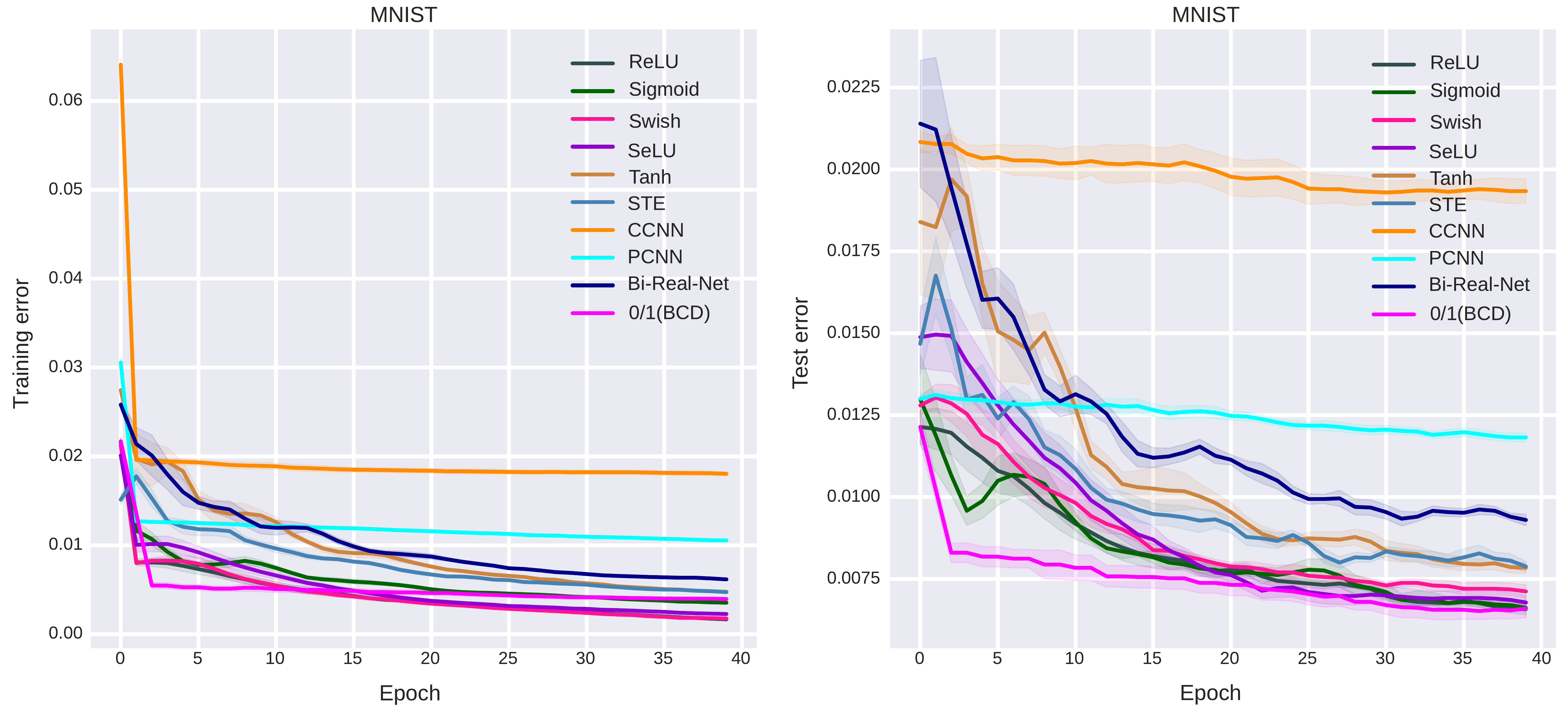}\\  \vspace{1mm}
\includegraphics[width=1\textwidth]{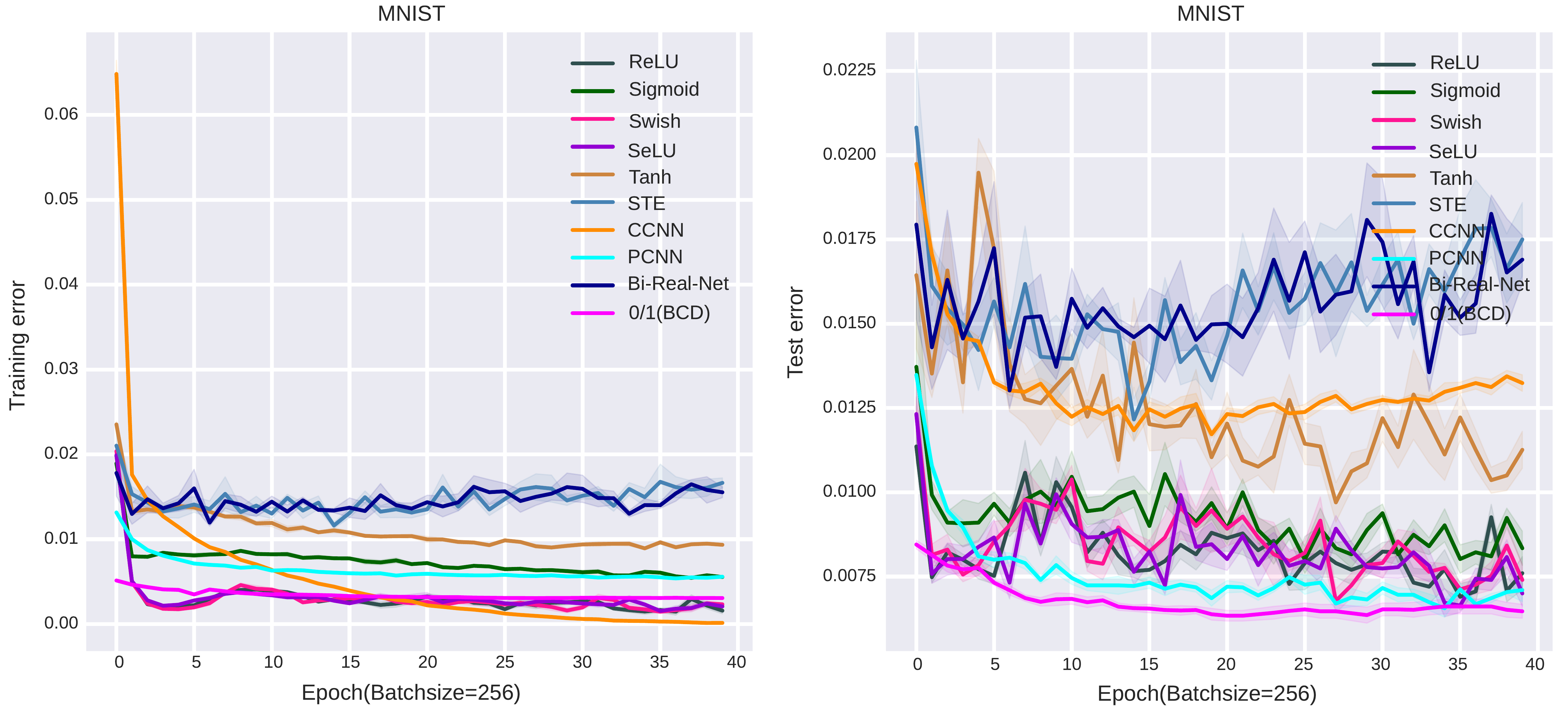}\\  
\caption{Performance on classifying MNIST: The top two are trained with full batches, the bottom two are trained with small batches.}
\label{fig:MNIST} 
\end{figure*}

The implementations for BCD are as follows: We initialize the weights in $\CW^0$ as small random Gaussian values without any pretreatments. Parameters are set as $\pi = 10^{-7}$, $\gamma = 10^{-8}$, $\tau = 10^{-6}$, and $\lambda=0.052$. The maximum number of iterations is $K=35.$ We shall emphasize that the selections of these parameters would not impact BCD significantly.  For Algorithm \ref{ses}, we fix $\beta=0.00072$ and choose $L=1$ for the purpose of accelerating the computation.


\subsection{Benchmark methods}
As mentioned above, we will compare 0/1 DNNs with DNNs with seven different activation functions. They are the Sign function, Binarization function, ReLU,   Sigmoid, Swish (with $\kappa=0.7$ used in this paper), SELU, and Tanh. The first two are 1-bit activation functions and the last five are continuous activation functions as shown in Fig. \ref{figureactivation}. Their analytical expressions are given as follows.
\begin{eqnarray*}
 \arraycolsep=0.5pt\def\arraystretch{1.25}
~\begin{array}{lll}
\sigma_{\rm Sign}(t)& =&  \left\{
\begin{array}{ll}
-1 ,& {\rm if~} t<0,\\
1, & {\rm if~} t\geq 0,
\end{array}\right.~~
\sigma_{\rm Binarization}(t) =  \left\{
\begin{array}{ll}
0 ,& {\rm if~} t<0,\\
1, & {\rm if~} t\geq 0,
\end{array}\right.\\
 \sigma_{\rm ReLU}(t) &=& \max\{t,0\},\qquad~
 \sigma_{\rm Sigmoid}(t) =  {1}/({1+e^{-x}}),\\
 \sigma_{\rm Swish}(t) &=& t \sigma_{\rm Sigmoid}(\kappa t), ~~\kappa\in(0,1),\\
\sigma_{\rm SELU}(t) &=& \left\{
\begin{array}{ll}
1.0507 \times 1.6732  (e^t-1),~& {\rm if~} t\leq0,\\
1.0507 t, & {\rm if~} t>0,
\end{array}\right.\\
 \sigma_{\rm Tanh}(t) &= &(e^{t}-e^{-t})/(e^{t}+e^{-t}). 
\end{array}
\end{eqnarray*}
{As mentioned in Section \ref{Section-relatedlectures}, BNNs take 1-bit activation functions and are challenging to train directly. Popular training approaches are based on approximations. We select four state-of-the-art methods: STE \cite{courbariaux2015}, Bi-Real Net \cite{Martinez-Yang2020}, CCNN \cite{Xu-Cheung2019}, and PCNN \cite{Gu-Zhang2019}. Overall, we may have 9 types of DNNs  and one our 0/1 NDDs. To ensure a fair comparison, we endeavor to maintain similar setups across them. Hence,   we continue to use floating point values  instead of using 1-bit values as weights for BNNs.  The detailed architectures of the networks and the optimizers are presented in Table \ref{setupdiff}.}
\begin{table}[!th]
	\renewcommand{\arraystretch}{1.25}\addtolength{\tabcolsep}{-3.5pt}
\centering
\caption{Setup for different solvers. \label{setupdiff}}
\begin{tabular}{c|c|c|c}\hline
Activation &$0/1$&Continuous& 1-bit\\\hline 
\multirow{2}{*}{Networks}&\multirow{2}{*}{0/1(BCD)}&ReLU, Sigmoid,& STE, CCNN, PCNN \\
&&Swish, SELU, Tanh& Bi-Real-Net\\\hline
MNIST$\&$&{hdmax+MSE+}&\multicolumn{2}{c}{sfmax+CE+}\\
FashionMNIST&F3NN+BCD&\multicolumn{2}{c}{F3NN+Adam}\\\hline
\multirow{2}{*}{Cifar10} &hdmax+MSE+&\multicolumn{2}{c}{\multirow{3}{*}{sfmax+CE+}}\\
&Resnet18+BCD&\multicolumn{2}{c}{\multirow{3}{*}{Resnet18+Adam}}\\\cline{1-2}
\multirow{2}{*}{Cifar100} &sfmax+CE+& \multicolumn{2}{c}{}\\
&Resnet18+BCD& \multicolumn{2}{c}{}\\\hline
\end{tabular}
\end{table}

{For instance, in cases of ReLU and Cifar10, `sfmax+CE+Resnet18+Adam' means that the network  adopts the   Resnet18 architecture, uses ReLU activation functions in the hidden layers and softmax in the output layer, and is trained by the famous optimizer Adam.  We  opt for  Adam because it not only is one of the most popular gradient-driven algorithms at present but also sets fixed parameters to avoid poor performance from improper parameter adjustments. Moreover, for more efficient training, we preprocess the data consistently across all networks: pre-train data by a small batch size at the beginning of several epochs (one epoch for MNIST, two epochs for FashionMNIST). }

\begin{figure*}[!th]
\centering
\includegraphics[width=1\textwidth]{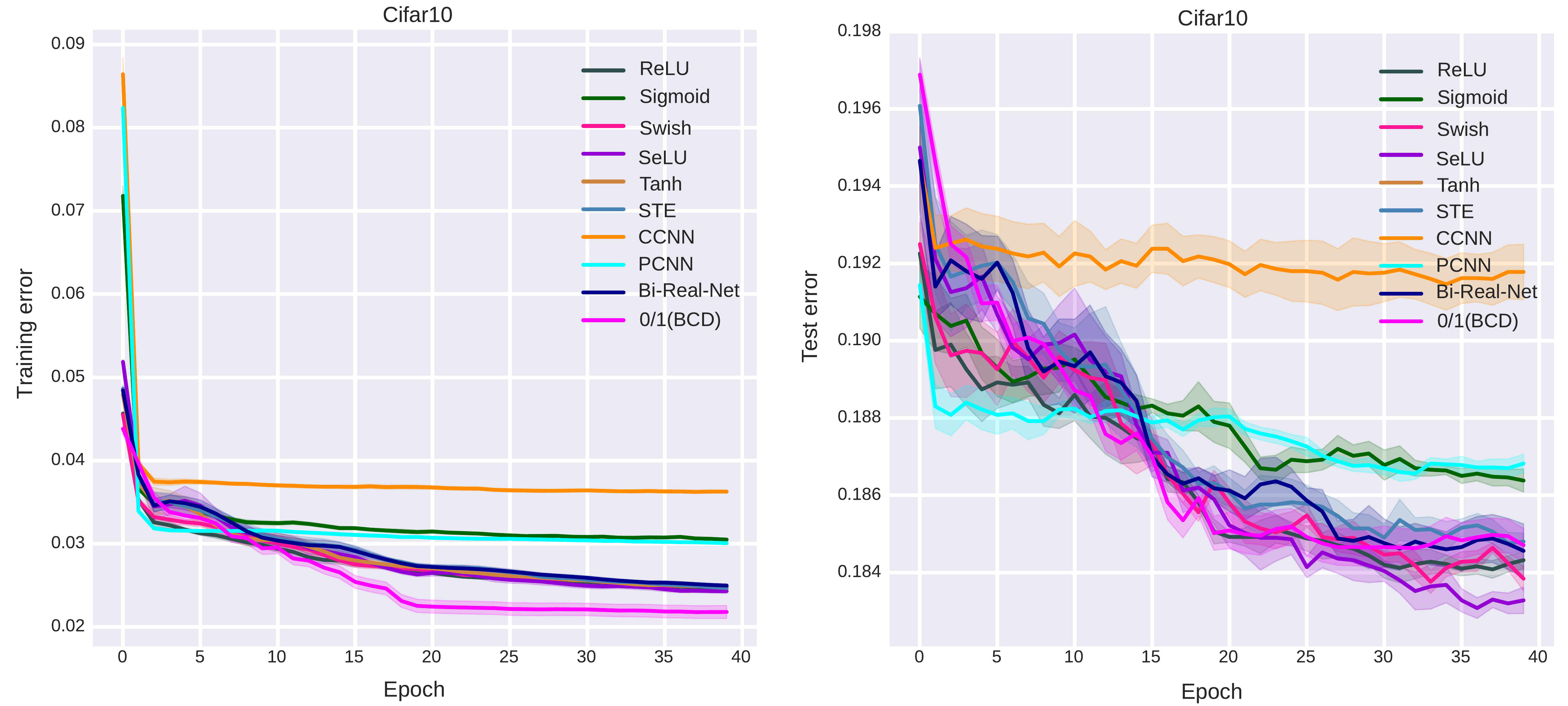}\vspace{-2mm}
\caption{Performance on classifying  Cifar10.}
\label{fig:Cifar10}
\end{figure*}

\begin{figure*}[!th]
\centering
\includegraphics[width=1\textwidth]{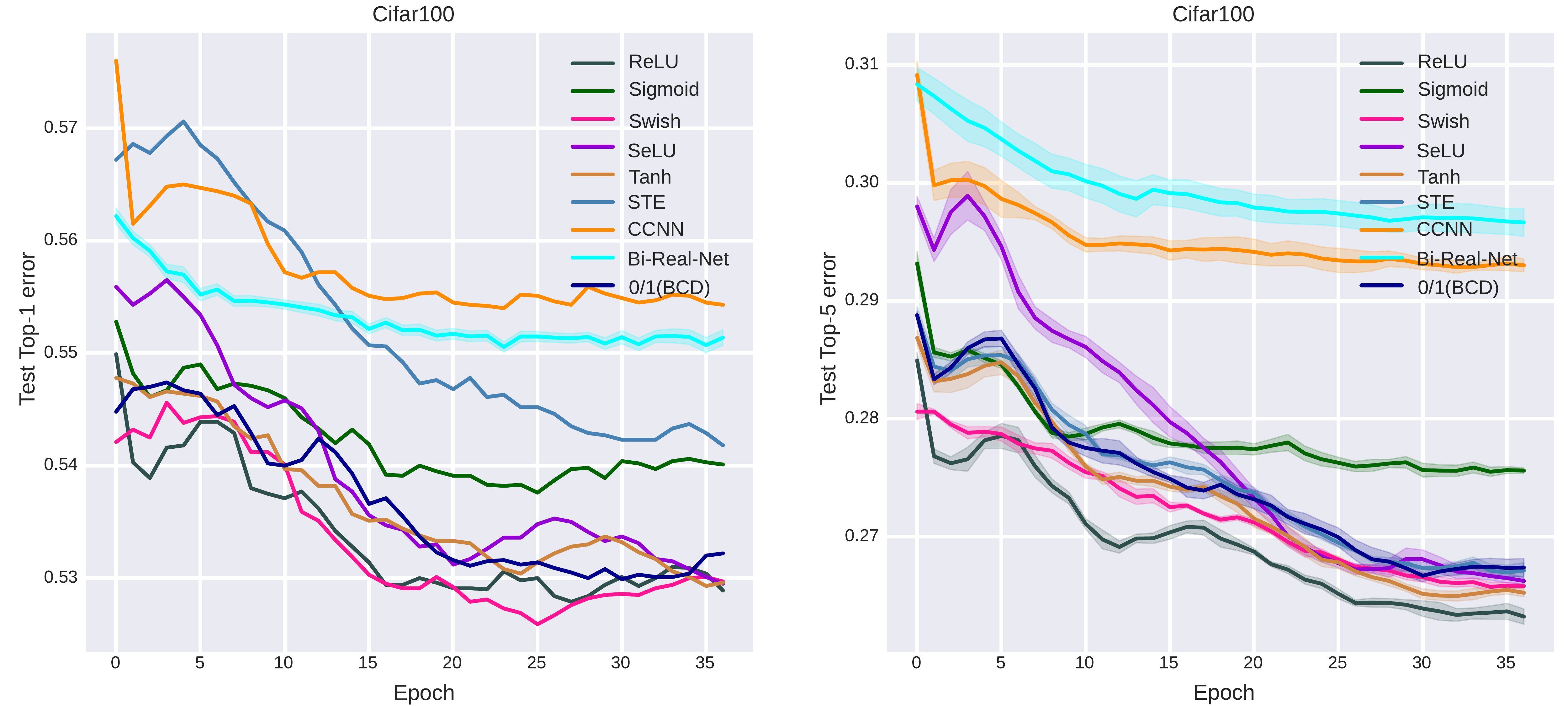}
\vspace{-5mm}
\caption{Performance on classifying Cifar100.}
\label{fig:Cifar100Top5}\vspace{-2mm}
\end{figure*}

We will report the following five metrics
\begin{center}
(Tr. error,~~Te. error,~~Fil.~num.,~~Par.~num., FLOPs.)
\end{center}
to illustrate the performance of all methods. The first four metrics represent the error on the training dataset, the error on the testing dataset, the number of nodes/filters in hidden layers, and the number of network parameters, respectively. The FLOPs stands for the floating-point operations per second. { We point out that Tr. error and Te error are  Top-1 errors for all datasets. However, we also report the Top-5 test errors for Cifar100. To reduce the influence of the initial point on the network training, we test each network five times with five randomly initialized points. Training and test errors of 10 methods are shown in  Figs. \ref{fig:MNIST}-\ref{fig:Cifar100Top5}, where solid curves and shaded regions respectively represent the mean and standard deviation of five trials.}

{
\subsection{Numerical results}
\subsubsection{Training and testing accuracy} According to Tables \ref{com-9-algs1} and  \ref{com-9-algs2}, it is evident that ReLU yields the lowest training errors when classifying MNIST (with a full batch size), MNIST (with Batchsize=256), Cifar10, and Cifar100, while Swish obtains  the best one for classifying FashionMNIST. In contrast, our algorithm 0/1(BCD) generates relatively competitive training errors and the best testing errors for most of the datasets. 
  
  \begin{figure*}[!t]
\centering
\includegraphics[width=1\textwidth]{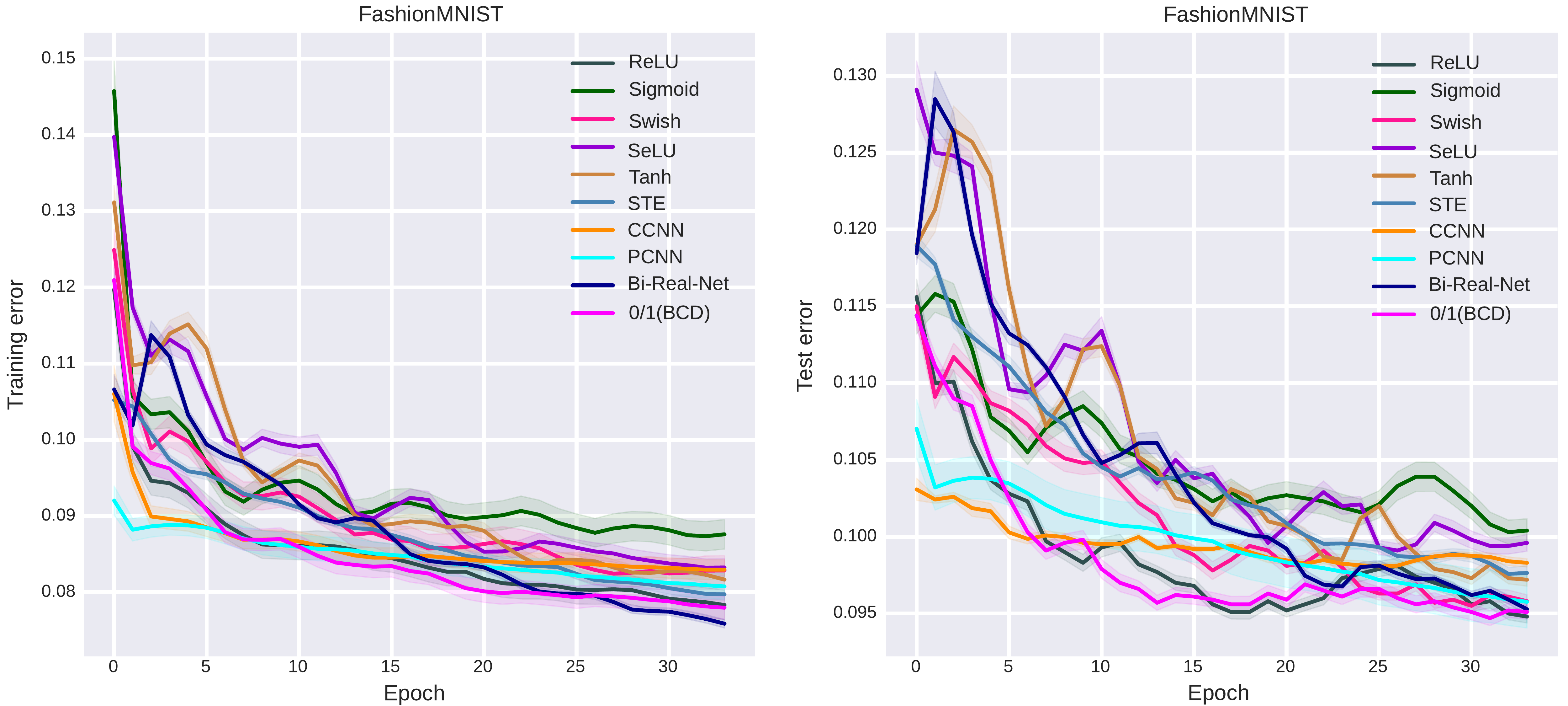}
\caption{Performance on classifying FashionMNIST.}
\label{fig:Fashion}\vspace{-3mm}
\end{figure*}

\begin{table*}[!tb] \renewcommand{\arraystretch}{1.0}\addtolength{\tabcolsep}{-4pt}
\centering
\caption{{Comparison of nine methods.}}\label{com-9-algs1}
\begin{tabular}{l |ccccc| ccccc |ccccc |ccccc}\hline
{~} & Tr.   & Te.  &  Fil. &  Par. &  FLOPs & Tr.   & Te.  &  Fil. &  Par. &  FLOPs& Tr.   & Te.  &  Fil. &  Par. &  FLOPs& Tr.   & Te.  &  Fil. &  Par. &  FLOPs\\ 
{~} & error &  error &  num. &  num. &  & error &  error &  num. &  num. &  & error &  error &  num. &  num. &  & error &  error &  num. &  num. &  \\
{~} & ($\%$) &  ($\%$) &   &    &  (M) & ($\%$) &  ($\%$) &    &    &  (M)& ($\%$) &  ($\%$) &    &    &  (M)& ($\%$) &  ($\%$) &   &    &  (M)\\\hline 
&\multicolumn{5}{c|}{MNIST}&\multicolumn{5}{c|}{MNIST (Batchsize=256)}
&\multicolumn{5}{c|}{FashionMNIST}
&\multicolumn{5}{c}{Cifar10}\\
\hline
ReLU   &{\bf 0.166}  & 0.657   &  4000 & 5172k & 0.558&{\bf 0.146}  & 0.690   &  4000 & 5172k & 0.558&7.831     & 9.480     & 4000       & 5172k  & 0.558& 2.478 & 18.432   &  3000 & 14.21M & 223.4\\
Sigmoid   & 0.353  & 0.662&  4000 & 5172k & 0.558& 0.545  & 0.800&  4000 & 5172k & 0.558& 8.760  & 10.04   & 4000     & 5172k & 0.558 &3.050   & 18.638&  3000 & 14.21M & 223.4\\
Swish      & 0.175  & 0.712& 4000 & 5172k & 0.558 & {\bf 0.146}  & 0.678& 4000 & 5172k & 0.558&8.285  & 9.580    & 4000     & 5172k & 0.558& 2.486  &{\bf 18.384}& 3000 & 14.21M & 223.4\\
SELU          &0.226  & 0.678& 4000 & 5172k& 0.558 &0.154  & 0.664& 4000 & 5172k& 0.558& 8.325  & 9.960   & 4000   & 5172k & 0.558& 2.430 & 18.328 & 3000 & 14.21M& 223.4\\
Tanh         &0.475  & 0.784 & 4000 & 5172k&0.558 &0.892  & 0.970 & 4000 & 5172k&0.558& 8.165  & 9.720     & 4000   & 5172k & 0.558&2.488   & 18.430 & 3000 & 14.21M&223.4\\
STE         &0.475  & 0.788 & 4000 & 5172k& 0.382&1.161  & 1.211 & 4000 & 5172k& 0.382&7.797  & 9.764      & 4000   & 5172k& 0.382&2.465  & 18.480 & 3000 & 14.21M& 221.7\\
CCNN         &1.852  & 1.891 & 4000 & 5172k& 0.382&1.135  & 1.171 & 4000 & 5172k& 0.382&8.344  & 9.785 & 4000 & 5172k& 0.382&3.626  & 19.178& 3000 & 14.21M& 221.7\\
PCNN        &1.031  & 1.203 & 4000 & 5172k& 0.382&0.538  & 0.656 & 4000 & 5172k& 0.382&7.812  & 9.496 & 4000 & 5172k& 0.382&3.007  & 18.681& 3000 & 14.21M& 221.7\\  
Bi-Real-Net     &0.514  & 0.879 & 4000 & 5172k& 0.382 &1.192  & 1.306 & 4000 & 5172k& 0.382&{\bf 7.632}  & 9.462 & 4000 & 5172k& 0.382&2.497 & 18.456 & 3000 & 14.21M& 221.7\\
0/1(BCD) & 0.395  & {\bf 0.652}   &  {\bf 3800}  & {\bf 4294k}  & {\bf 0.363}& 0.306  & {\bf 0.634}   &  {\bf 3800}  & {\bf 4294k}  & {\bf 0.363}&7.780   & {\bf 9.451}   & {\bf 3800}        & {\bf 4294k}& {\bf 0.363}&{\bf 2.179}    & 18.470  &  {\bf 2340}  & {\bf 14.20M}  & {\bf 221.4}\\   
\hline
\end{tabular}
\end{table*}
 \subsubsection{FLOPs and Network scale}  It can be clearly seen from Tables \ref{com-9-algs1} and  \ref{com-9-algs2} that 0/1(BCD) enables a more efficient network with lowest FLOPs followed by STE, CCNN, PCNN, and Bi-Real-Net. This is because the binary operation is capable of accelerating the computation and the $\ell_{2,0}$-regularization can sparsify the network. Consequently, the number of active nodes in the trained network is lower compared to other networks, e.g., Fil. num.$=3800$ v.s. Fil. num.$=4000$.}

  \subsubsection{Effect of the initial process} As presented in Fig. \ref{fig:MNIST}, the initial process impacts on all methods to some extend. If there is no such an initial process (corresponding to  0 epochs), then all methods obtain relatively high training error. However, with the increasing number of epochs, the training and testing errors decline dramatically at first and then tend to be steady. 
  
   \begin{table}[H] \renewcommand{\arraystretch}{1}\addtolength{\tabcolsep}{0.3pt}
\centering
\caption{{Comparison of eight methods for Cifar100.}}\label{com-9-algs2}
\begin{tabular}{lccccc}\hline
{~} & Top-1 & Top-5 & {Fil.  num.} & {Par. num.}& {FLOPs}\\ \hline
\multicolumn{6}{c}{Cifar100}\\\hline
ReLU   &{\bf 52.934$\%$}  & {\bf 26.321$\%$}   &  4200 & 16.54M & 225.6M\\
Sigmoid   & 54.019$\%$  & 27.544$\%$&  4200 & 16.54M  & 225.6M\\
Swish      & 52.982$\%$  & 26.572$\%$& 4200 & 16.54M  & 225.6M\\
SELU          &52.996$\%$  & 26.635$\%$& 4200 & 16.54M & 225.6M\\
Tanh         &52.993$\%$  & 26.523$\%$ & 4200 & 16.54M &225.6M\\
STE         &54.131$\%$  & 26.689$\%$ & 4200 & 16.54M & 183.0M\\
CCNN         &55.432$\%$  & 29.624$\%$ & 4200 & 16.54M & 183.0M\\
Bi-Real-Net     &55.167$\%$  & 29.239$\%$ & 4200 & 16.54M & 183.0M\\
0/1 & 53.189$\%$  & 26.691$\%$   &  {\bf 3990}  & {\bf 16.53M} & {\bf 182.9M}\\\hline
\end{tabular}
\end{table}

 \subsubsection{Robustness to the noise} 
\begin{figure}[H]
\centering
\includegraphics[scale=.3]{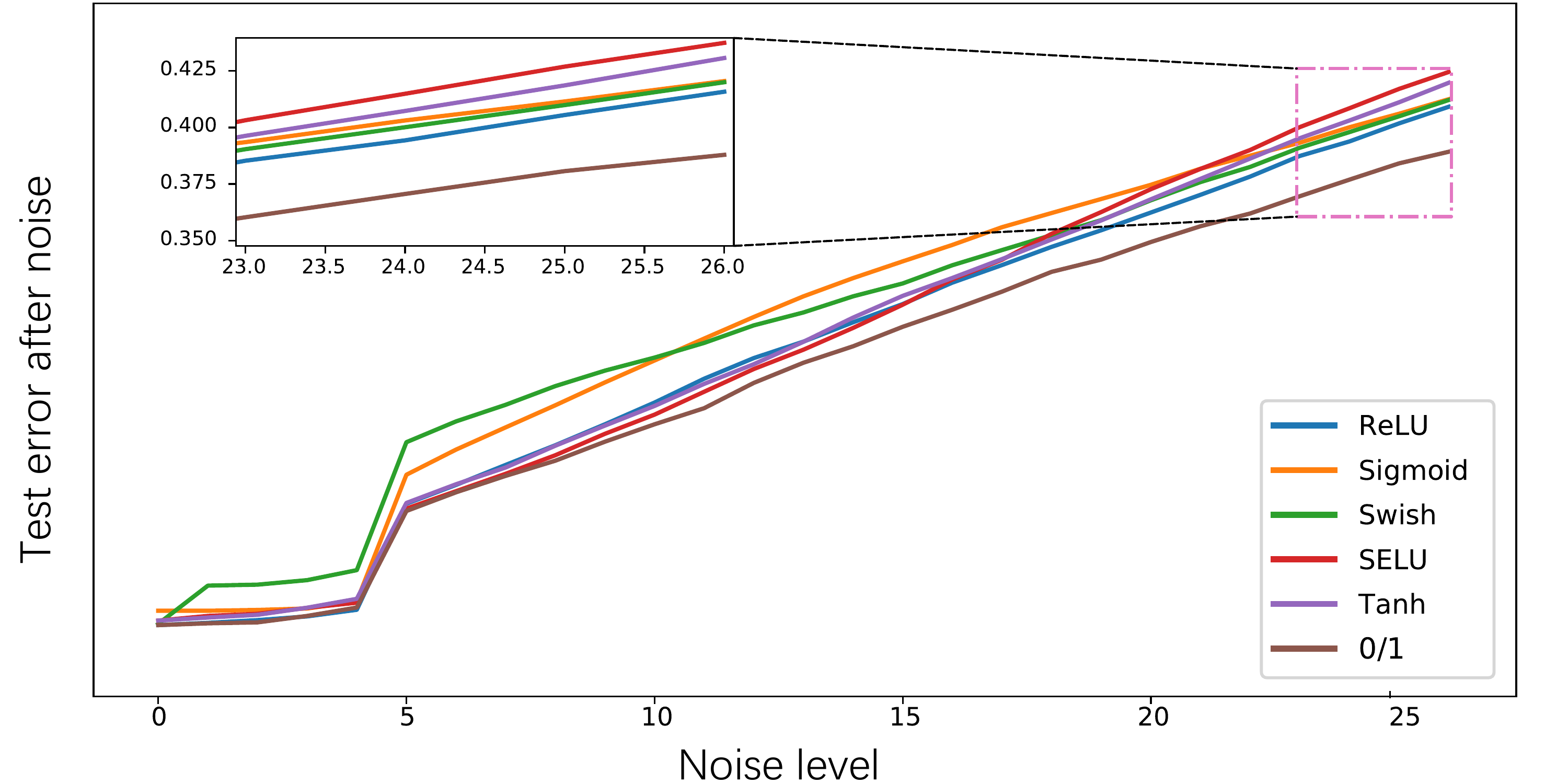}\vspace{-2mm}
\caption{Test error in different noise level.}
\label{fig:Robust}
\end{figure}

\begin{figure*}[!th]
\centering
\includegraphics[width=0.24\textwidth]{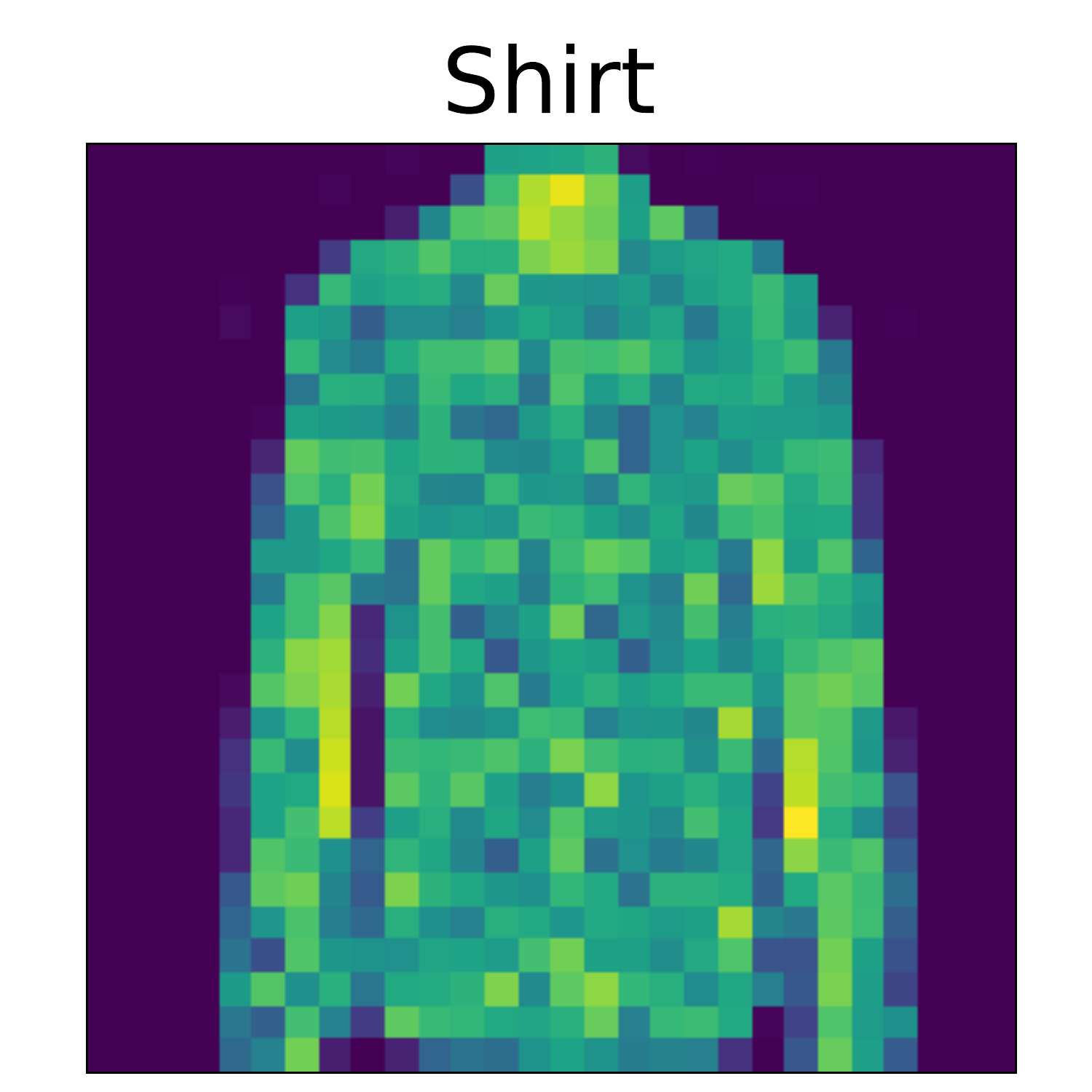}
\includegraphics[width=0.24\textwidth]{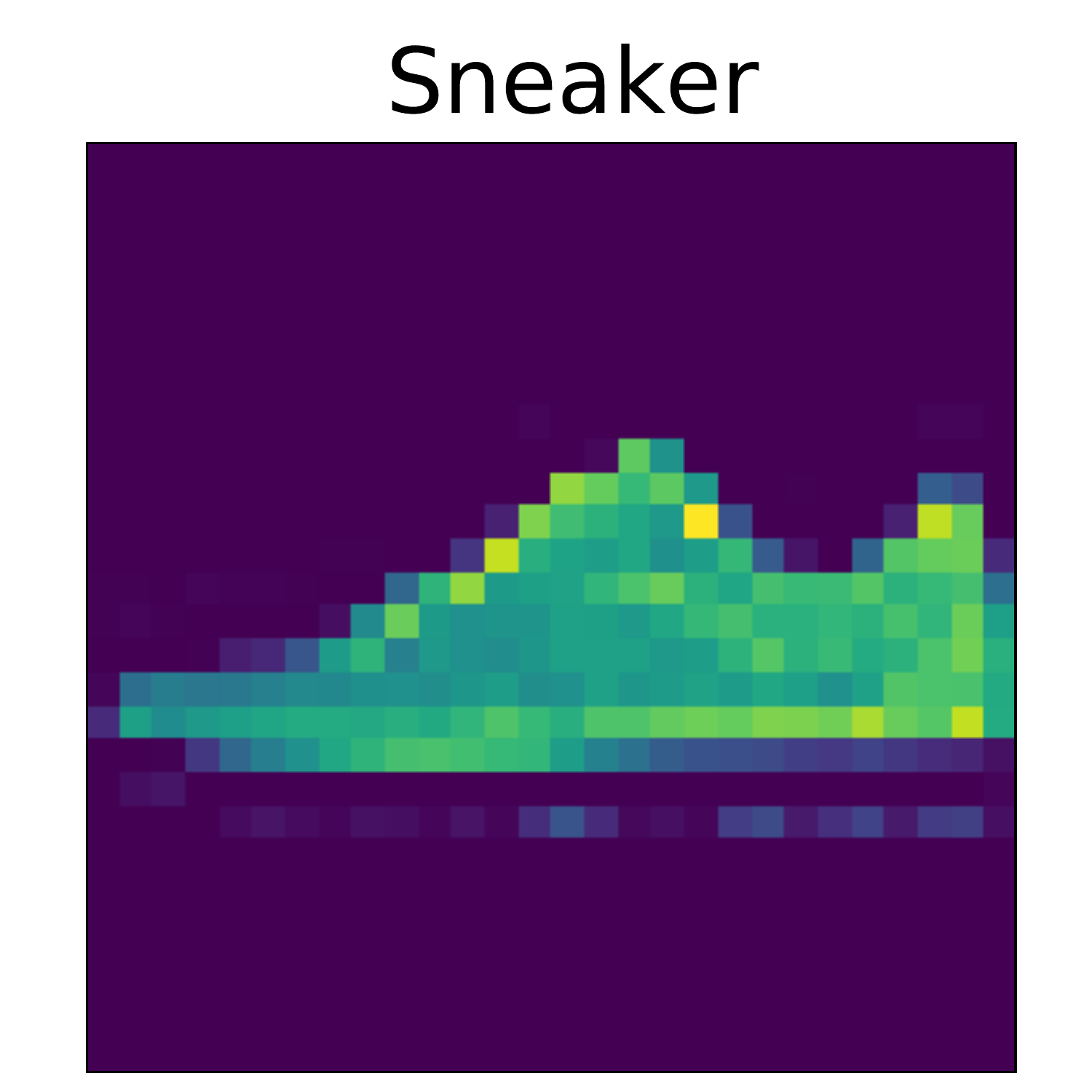}
\includegraphics[width=0.24\textwidth]{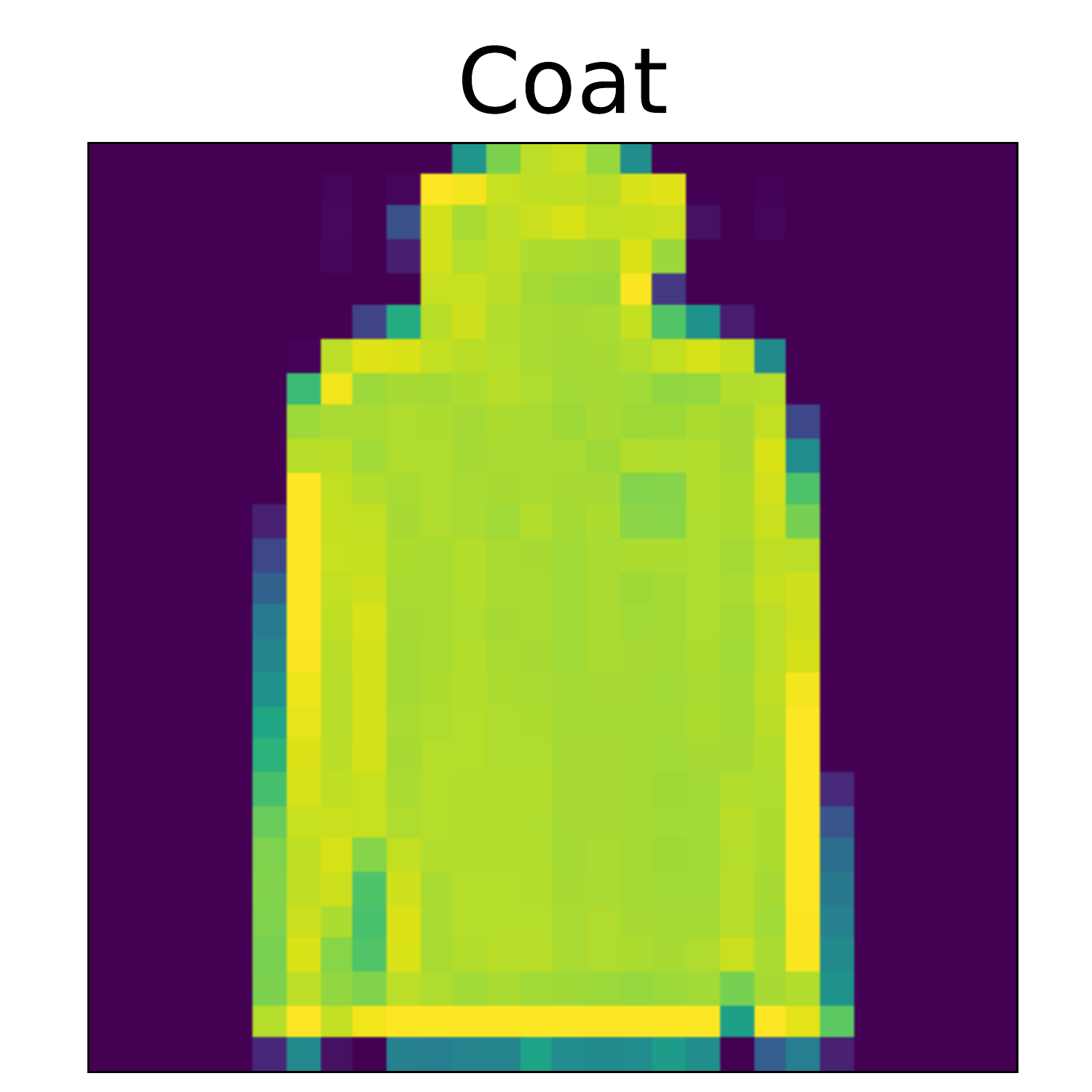}
\includegraphics[width=0.24\textwidth]{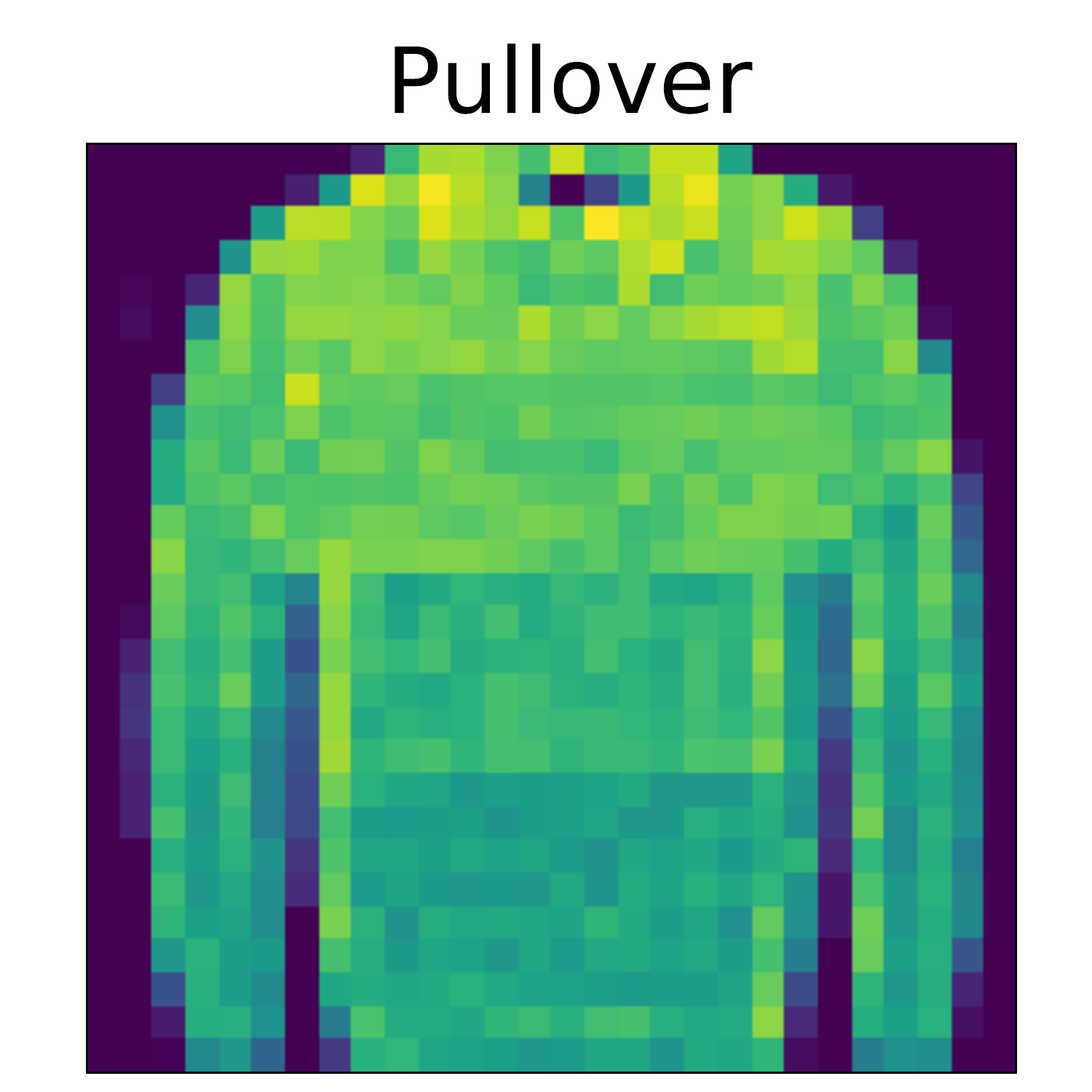}\\
\includegraphics[width=0.24\textwidth]{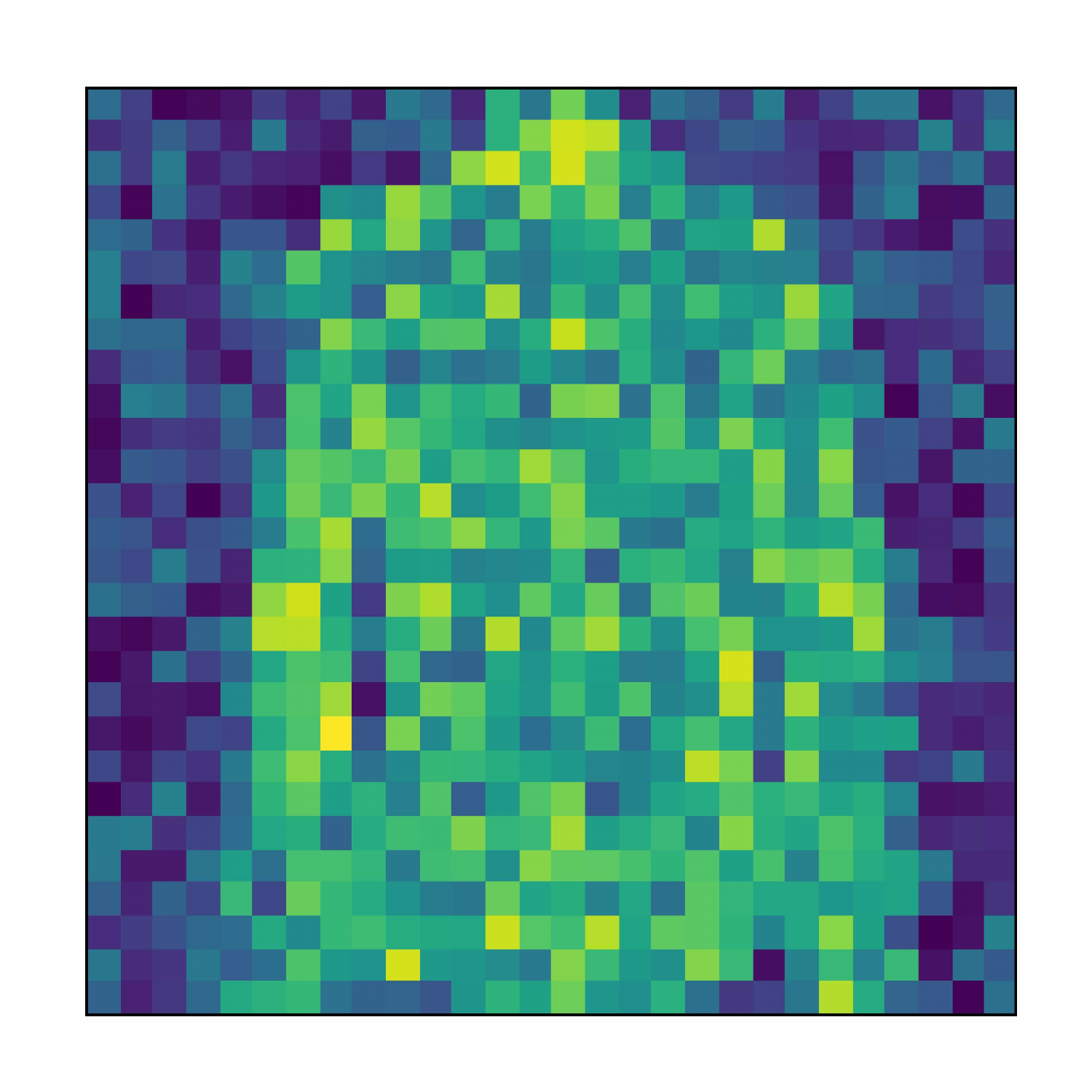}
\includegraphics[width=0.24\textwidth]{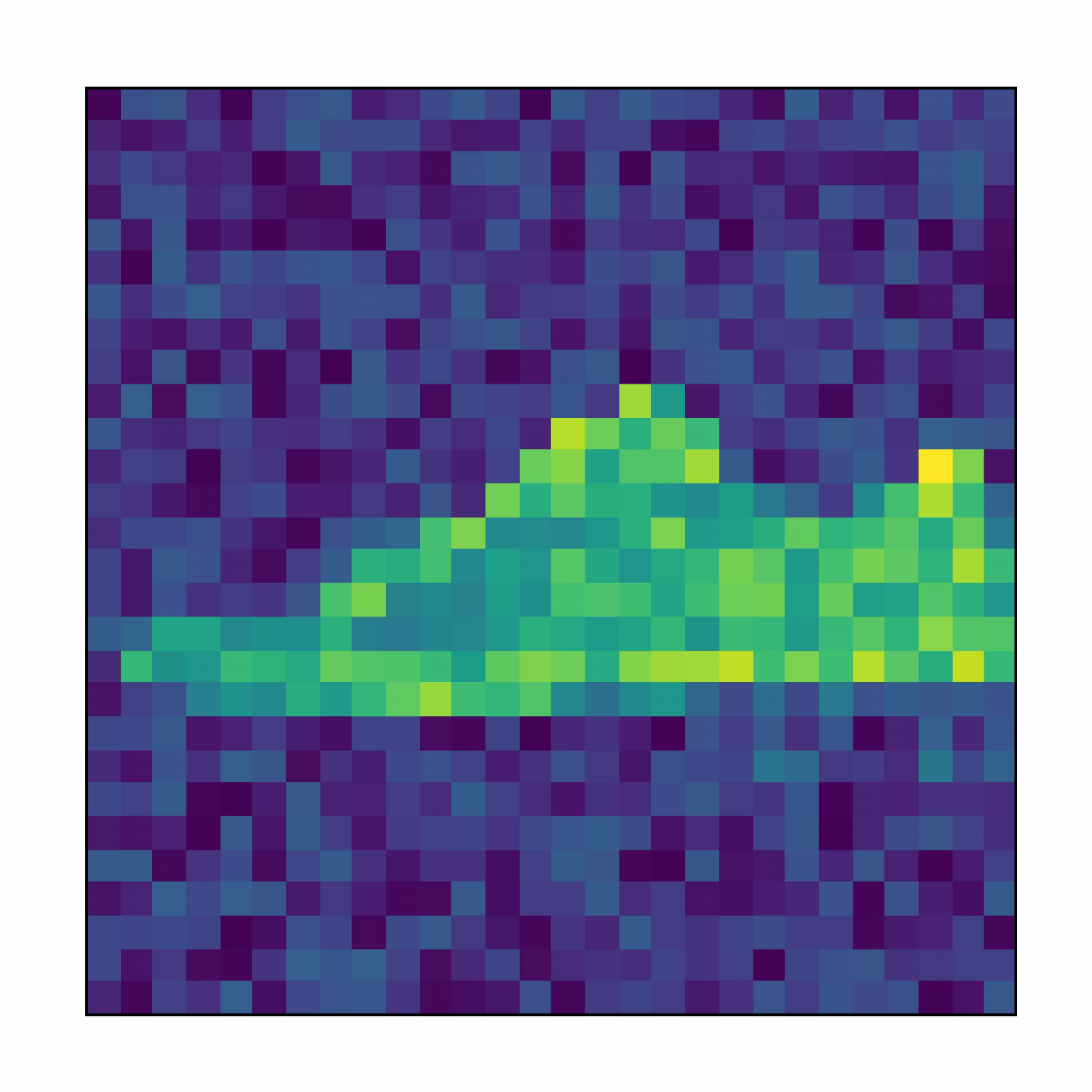}
\includegraphics[width=0.24\textwidth]{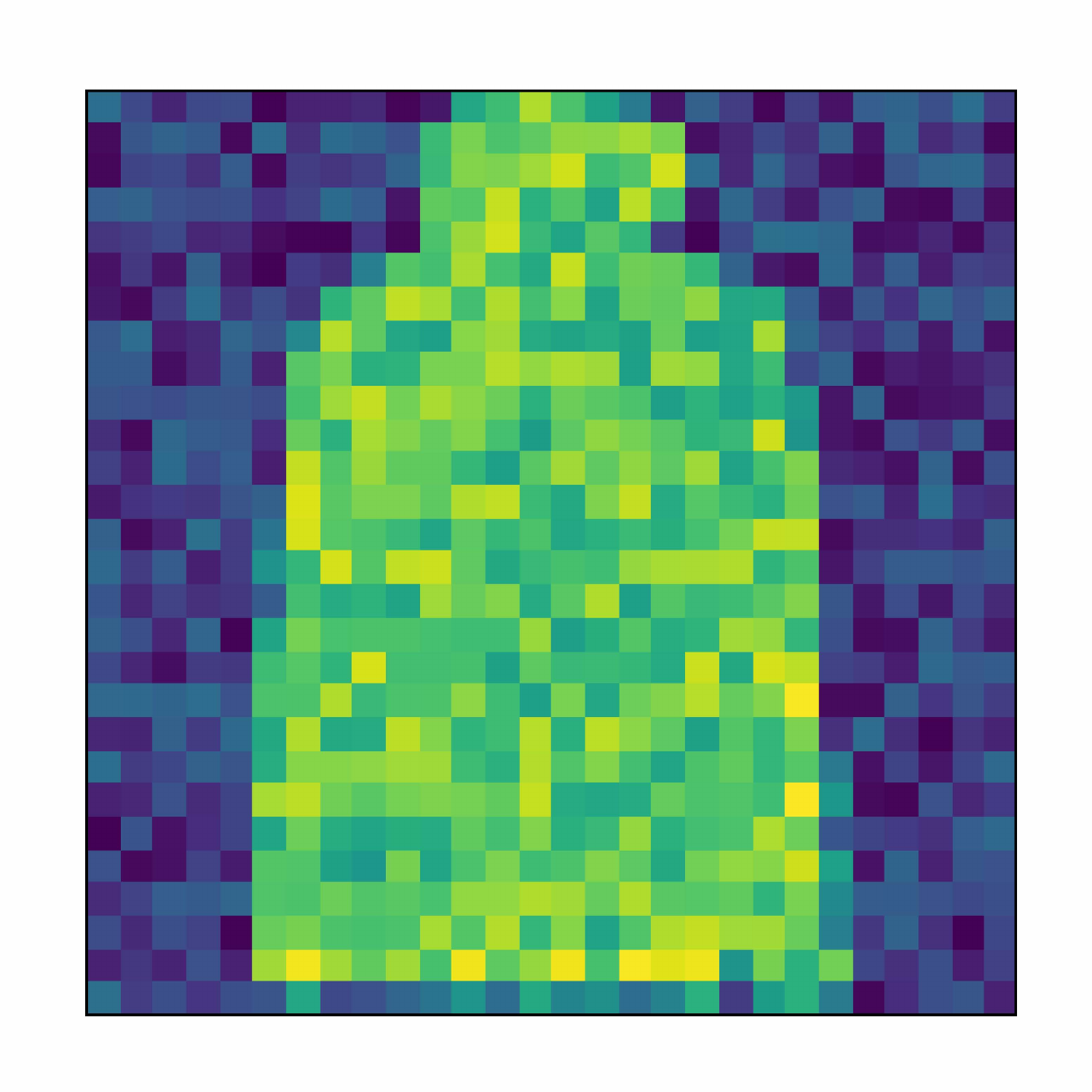}
\includegraphics[width=0.24\textwidth]{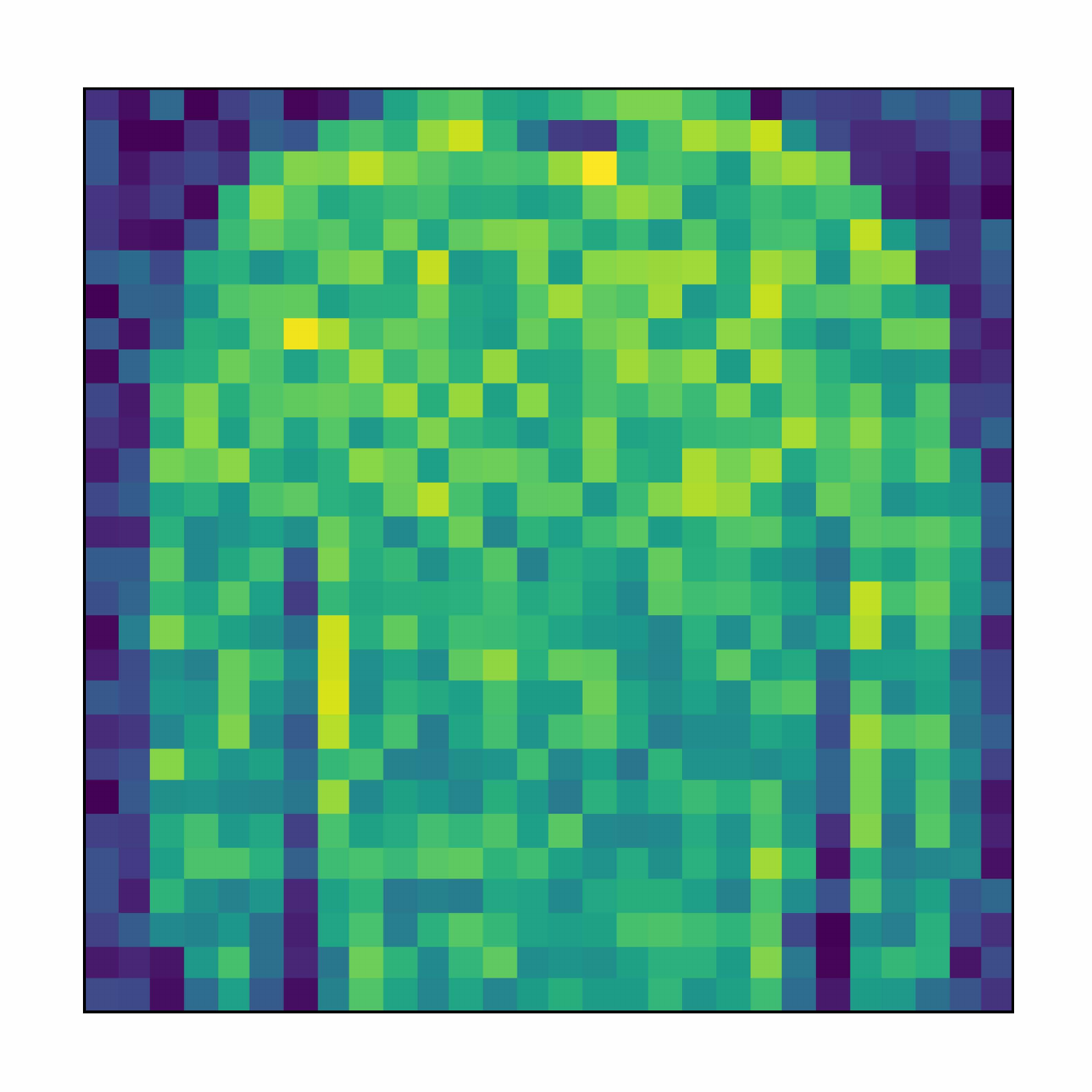}
\caption{Images in FashionMNIST. The first and second rows show  the original and noisy images.  
}
\label{rubfashion1}
\end{figure*}

\begin{figure*}[!th]
\centering
\includegraphics[width=0.475\textwidth]{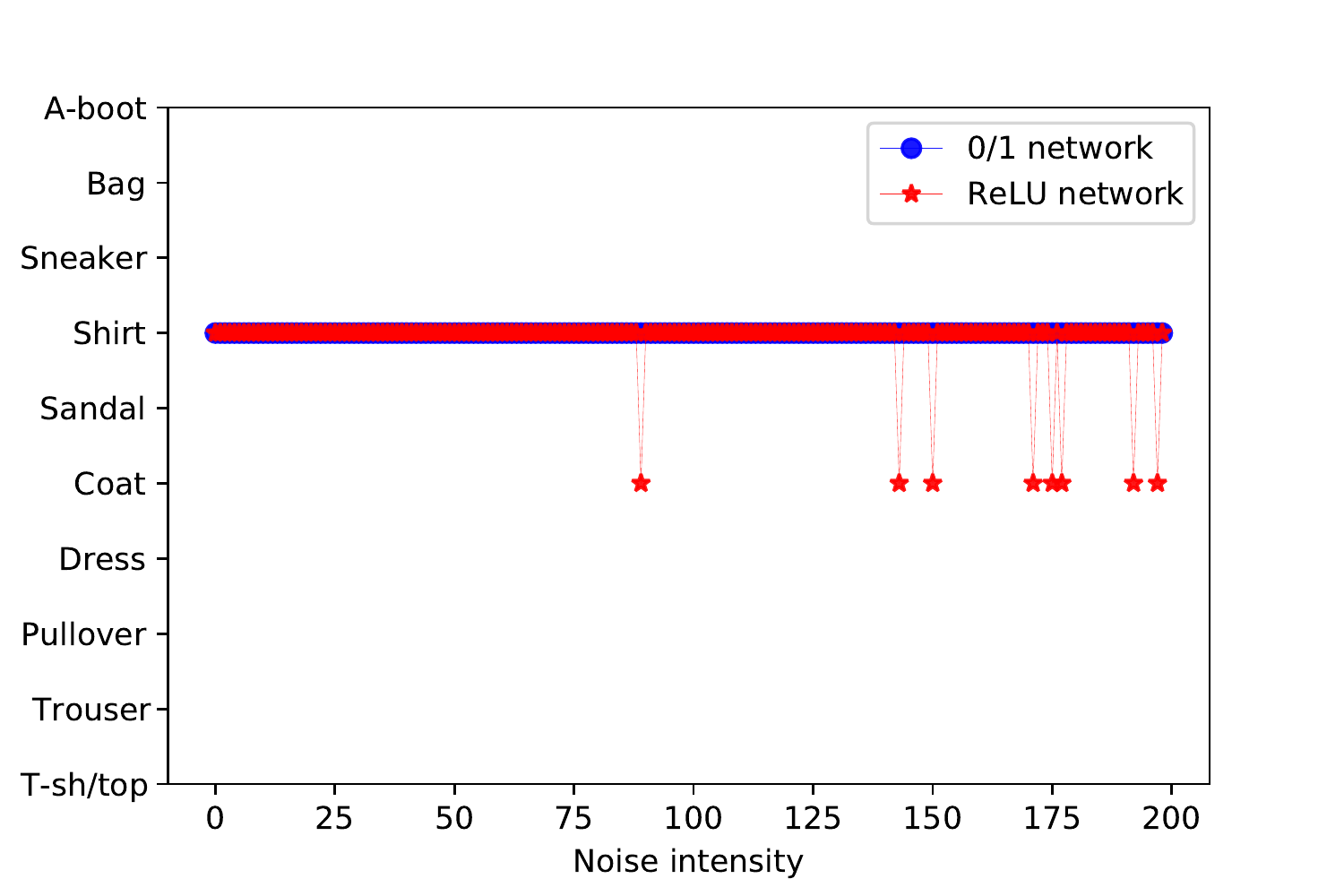}~~
\includegraphics[width=0.475\textwidth]{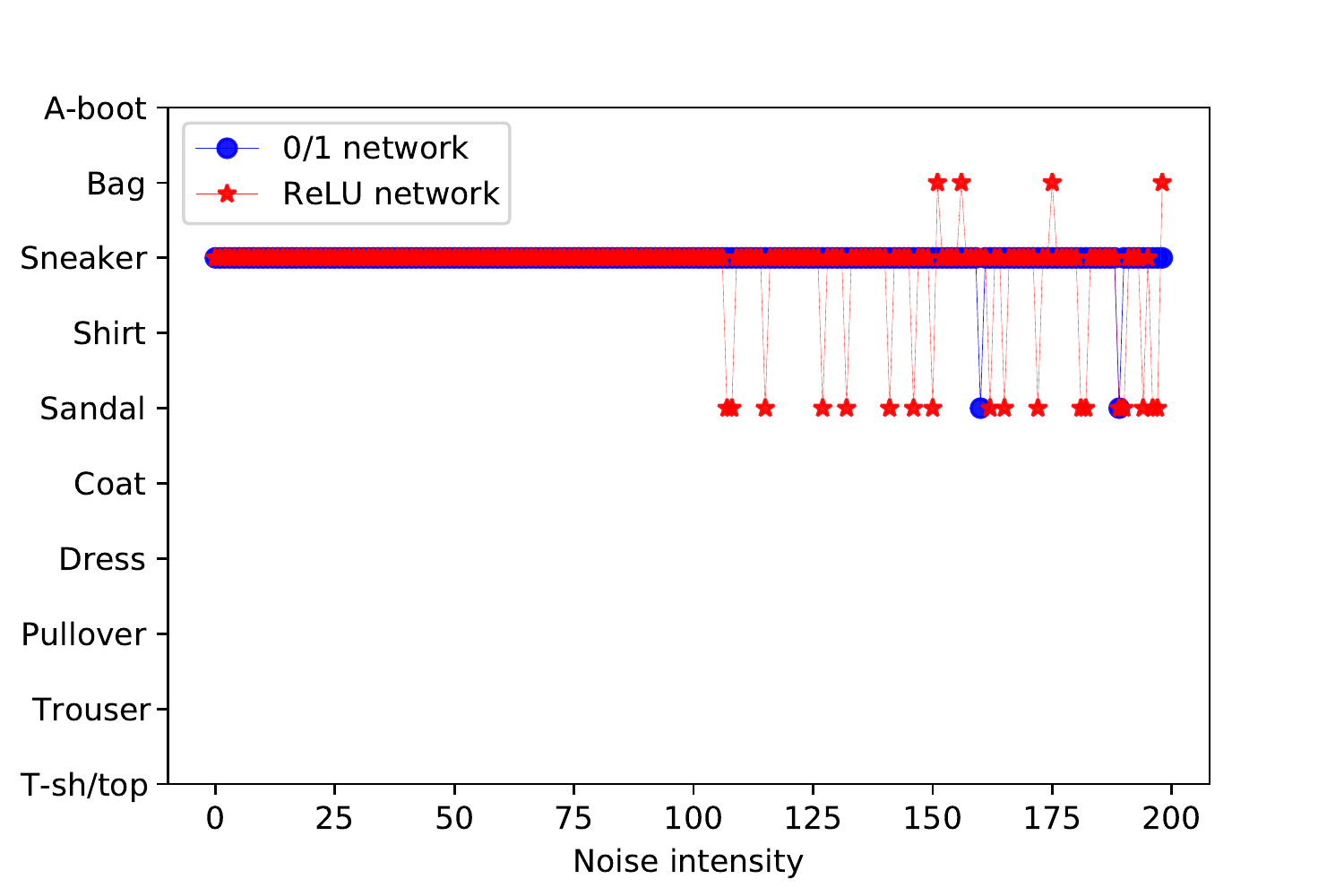}\\
\includegraphics[width=0.475\textwidth]{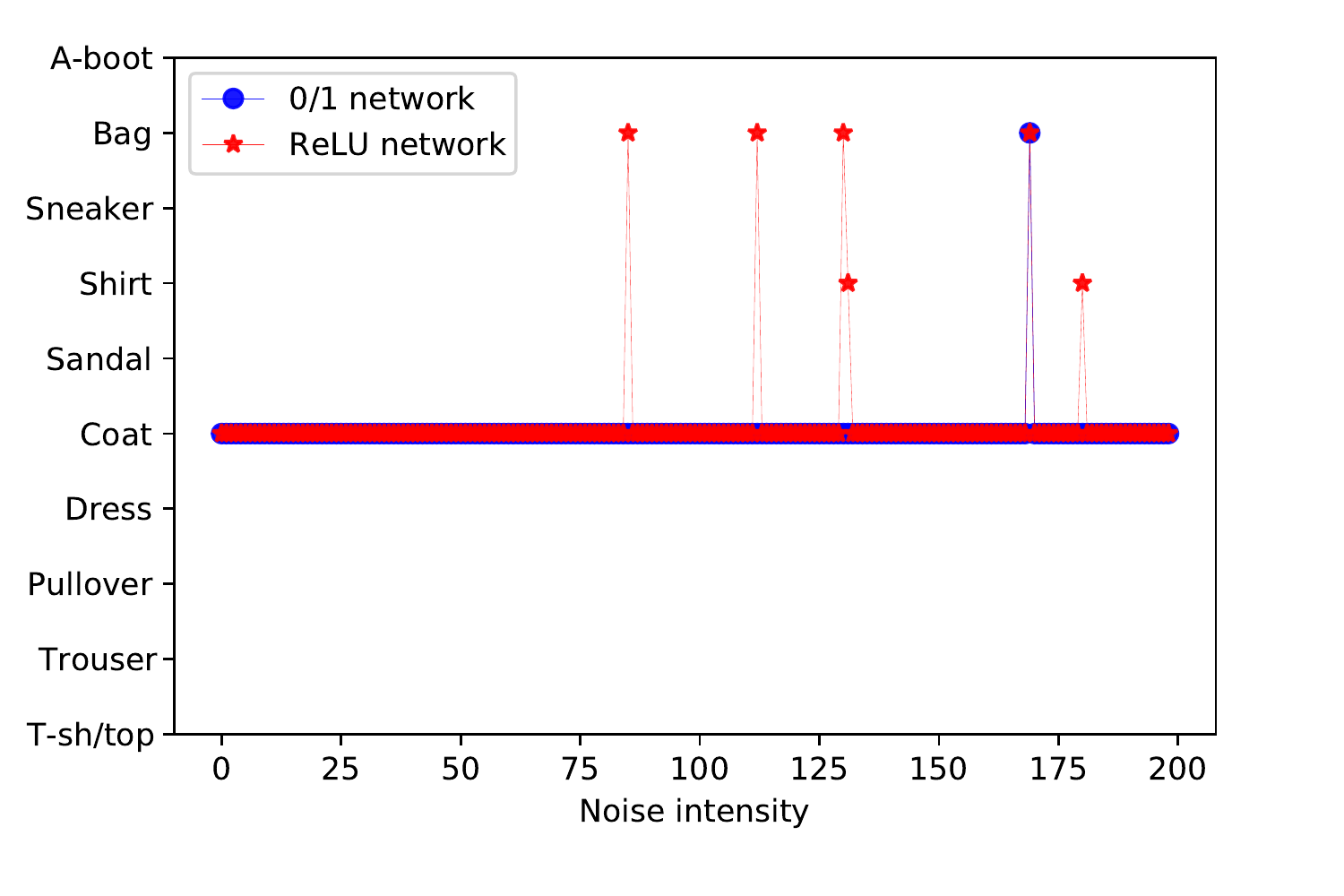}~~
\includegraphics[width=0.475\textwidth]{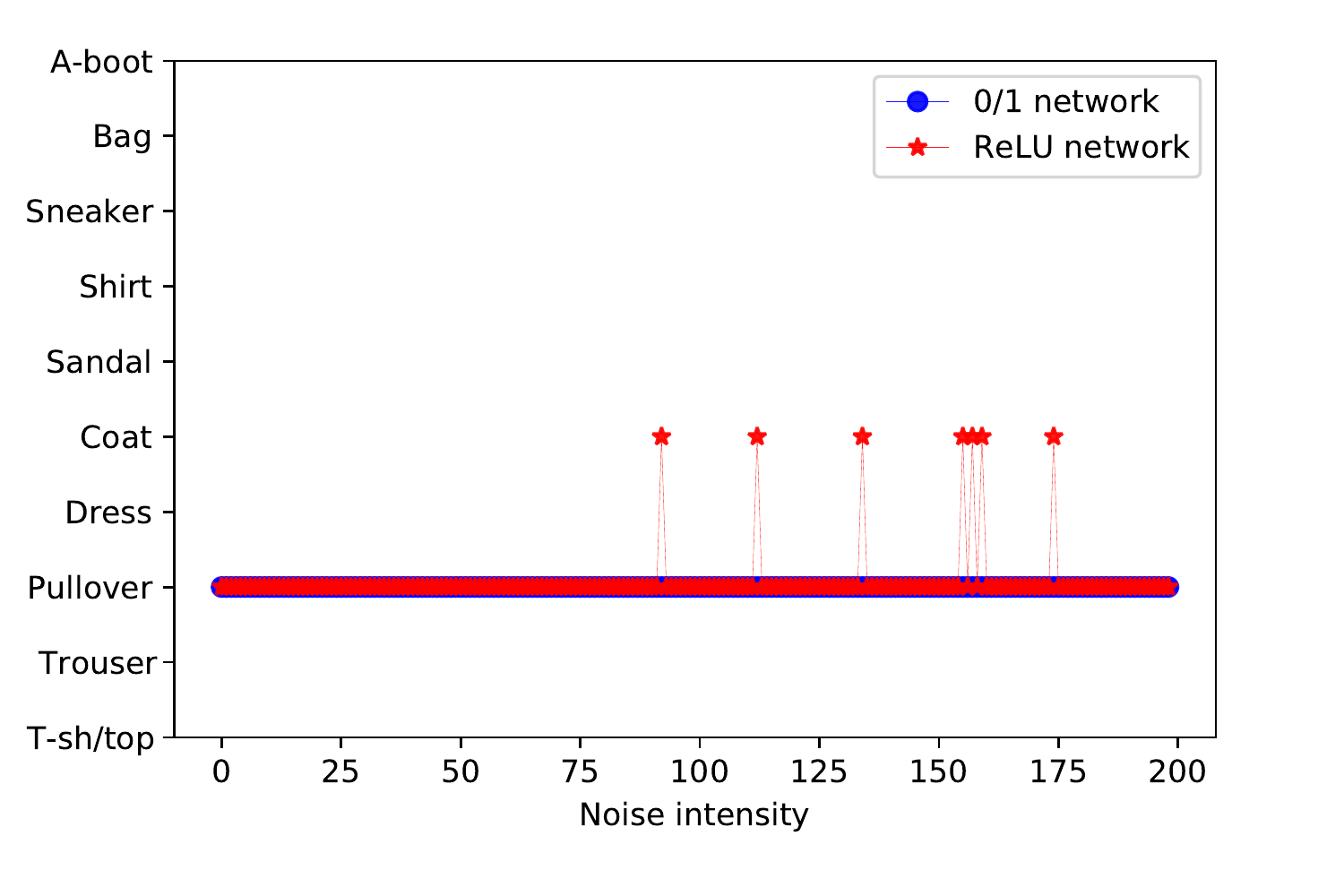}\vspace{-5mm}
\caption{Robustness to the noise.}
\label{rubfashion2}
\end{figure*}

Based on our experiment results,  0/1 DNNs are quite robust to adversarial perturbations to the noise. We add Gaussian noise to the original testing images of FashionMNIST, as shown in Fig.  \ref{rubfashion1}, and gradually increase the noise level. Then we use these noisy images to verify the stability of networks generated by 0/1 DNNs and ReLU DNNs. As present in Fig. \ref{rubfashion2}, for Shirt (in the top-left sub-figure) and Pullover (in the bottom-right sub-figure),  the noise levels do not impact the classifications by 0/1 DNNs. In contrast, ReLU DNNs frequently classify Shirts and Pullover as Coat for some higher noisy intensities.  Fig. \ref{fig:Robust} illustrates that the error of all test dataset with different noise level. From the figure, as the noise intensity increases, the test error of  0/1 DNNs is more stable than that for the other networks.

\section{Conclusion} \label{Section-conclusion}
The story of the 0/1 DNNs is historic, but there has been limited progress of research on developing algorithms specifically designed for 0/1 DNNs over the years. This is mainly due to the challenging of vanishing gradients of the 0/1 activation function that impedes the effectiveness of all backpropagation algorithms. Therefore, to train the 0/1 DNNs, we  cast a BCD algorithm that does not depend on the (sub)gradients heavily, thereby enabling training DNNs with other activation functions that may incur the issue of vanishing  gradients. {
Moreover, when using $\{0,1\}$ as weights, 0/1 DNNs can be deemed as a special case of BNNs. Exploring how to apply the framework of BCD to this scenario presents a promising topic of research.  Furthermore, in addition to the  image classification, we believe that the proposed 0/1 DNNs  could find utility in some other applications, such as medical diagnosis and automatic speech recognition. We leave this as the future research.}

\section*{Acknowledgements}
We acknowledge the support provided by the China Scholarship Council (CSC) during Dr. Hui Zhang's visit to Imperial College London. The authors extend their gratitude to Prof. Baochang Zhang and Dr. Jiaxin Gu from Beihang University for generously sharing the PCNN code and providing valuable insights regarding the numerical experiment.

\bibliographystyle{IEEEtran}

\newpage
\onecolumn
\appendices
 \section{Proofs of all theorems}
\subsection{Proof Lemma \ref{u-outlemma}}
\begin{proof}
Since  $\Omega:=\{\u: \phi_\u=\phi_\y\}={\rm argmin}_\u \|\y- (\u)_\hd\|^{2}$ and   $\b= {\rm argmin}_\u\|\u-\b\|^2$, two possible optimal solutions to problem \eqref{u-outproblem} are $\b$ or $\b+\epsilon \e_{\phi_\y}\in\Omega$, where $\epsilon> \b_{\max}-\b_{\phi_\y}=\Delta$. It is easy to show that $ \|\y-(\b+\epsilon \e_{\phi_\y})_\hd \|^2=0$ and
\begin{eqnarray*}
\begin{array}{lll}
\psi(\b)=\|\y-(\b)_\hd \|^2, \qquad \psi(\b+\epsilon \e_{\phi_\y})=\mu \epsilon^2.
\end{array}
\end{eqnarray*}
If $\mu \Delta^2 > \|\y-(\b)_\hd \|^2$, then the above formulas suffice to
\begin{eqnarray}
\begin{array}{lll}
\lim_{\epsilon\to\Delta^+} \psi(\b+\epsilon \e_{\phi_\y})=\mu\Delta^2 > \|\y-(\b)_\hd \|^2=
\psi(\b).
\end{array}
\end{eqnarray}
Similarly, we can derive the result for the other two cases.
\end{proof}

\subsection{Proof Lemma \ref{u_insolution}}
\begin{proof} One can find the possible solutions to \eqref{u-inproblem} are $u^*=0,b,\varepsilon$, and $-\varepsilon$, where $\varepsilon$ is a positive constant. Their objective function  values are summarized in Table \ref{obj-values}.

\begin{table}[H]
	\renewcommand{\arraystretch}{1.25}\addtolength{\tabcolsep}{10pt}
\centering
\caption{Four objective function  values, where `$\surd$' stands for the possible minimal value. \label{obj-values}}
\begin{tabular}{c llll}\hline
 &$\varphi (0)$& $\varphi (b)$& $\varphi (\varepsilon)$& $\varphi (-\varepsilon)$\\\hline
\multirow{2}{*}{ $b>0$} & $a^2+\rho  b^2$ & $(1-a)^2$ & $(1-a)^2+\rho  (\varepsilon-b)^2$ &   $a^2+\rho  (\varepsilon+b)^2$ \\
 &$\surd$&$\surd$&& \\\hline
\multirow{2}{*}{   $b=0$} & $a^2$ & $a^2$ & $(1-a)^2+\rho  \varepsilon^2$ &  $a^2+\rho  \varepsilon^2$ \\
 &$\surd$&$\surd$&$\surd$& \\\hline
\multirow{2}{*}{    $b<0$} & $a^2+\rho  b^2$ & $a^2$ & $(1-a)^2+\rho  (\varepsilon-b)^2$ &   $a^2+\rho  (\varepsilon+b)^2$ \\
 &&$\surd$&$\surd$& \\\hline
\end{tabular}
\end{table}\vspace{-3mm}
\noindent We only prove the result when $b<0$ as the other two cases are similar. If  $\varphi (\varepsilon)< \varphi (b)$, then  $2a-1 > \rho (\varepsilon-b)^2$. However, $\varphi (\varepsilon)$ is increasing with the rising of $\varepsilon>0$ due to $b<0$. Hence, we have $\varphi (u^*)=\lim_{\varepsilon \to0^+}\varphi (\varepsilon) = \varphi (\lim_{\varepsilon\to0^+} \varepsilon)$ and $u^*=\lim_{\varepsilon\to0^+} \varepsilon=0^+$.
\end{proof}
\subsection{Proof Theorem \ref{P-first-order-local}}
\begin{proof} For any $ \W\in {\mathbb N}(\W^*, \sqrt{\beta\lambda/(2m^2)})$, we denote two index sets as
  \begin{eqnarray*}
  \arraycolsep=1.4pt\def\arraystretch{1.25}
\ba{lll}
 \Xi:=\{s\in[m]:\W_{s:}\neq 0\},\qquad \Xi^*:=\{s\in[m]:[\W^*]_{s:}\neq 0\}.
 \ea
  \end{eqnarray*}
 We can prove that $\Xi^*\subseteq\Xi$. In fact, if there is  an index $t\in\Xi^*$ but  $t\notin\Xi$, then
 we obtain
  \begin{eqnarray*}
  \arraycolsep=1.4pt\def\arraystretch{1.25}
\ba{lll}
\sqrt{\beta\lambda/(2m^2)}   > \|\W -\W^*\| \geq \|\W^*_{t:}\|\geq \sqrt{2\beta\lambda},
 \ea
  \end{eqnarray*}
 where   the last inequality holds due to  \eqref{P-equivalent}. The above   contradiction implies that $\Xi^*\subseteq\Xi$. Therefore, $\|\W\|_{0,2}\geq \|\W^*\|_{0,2} $. If $\|\W\|_{0,2} = \|\W^*\|_{0,2} $, then $\Xi^*=\Xi$, which by the strong convexity of $ \Psi$ allows us to derive that
  \begin{eqnarray*}
  \arraycolsep=1.4pt\def\arraystretch{1.25}
\ba{lll}
&&\Psi(\W) + \lambda \|\W\|_{0,2} -( \Psi(\W^*) + \lambda \|\W^*\|_{0,2}) =\Psi(\W^*)   - \Psi(\W)  \\
 &\geq& \langle \nabla \Psi(\W^*), \W -\W^*  \rangle + \frac{\gamma}{2} \|\W -\W^*\|^2  \\
&=&\sum_{s\in \Xi^*}\langle 0, (\W -\W^*)_{s:}   \rangle + \sum_{s\notin \Xi^*}\langle (\nabla \Psi(\W^*))_{s:}, 0\rangle + \frac{\gamma}{2} \|\W -\W^*\|^2=\frac{\gamma}{2} \|\W -\W^*\|^2,
 \ea
 \end{eqnarray*}
 where the penultimate equality is due to \eqref{P-equivalent} and $\Xi^*=\Xi$,  showing \eqref{quadratic-growth-property}. If $\|\W\|_{0,2} > \|\W^*\|_{0,2} $, then $\|\W\|_{0,2} - \|\W^*\|_{0,2}\geq1$, which together with the strong convexity of $\Psi$  derives that
   \begin{eqnarray*}
  \arraycolsep=1.4pt\def\arraystretch{1.25}
\ba{lll}
&&\Psi(\W) + \lambda \|\W\|_{0,2} -( \Psi(\W^*) + \lambda \|\W^*\|_{0,2})  \\
&\geq &\Psi(\W)   - \Psi(\W^*) + \lambda \geq \langle \nabla \Psi(\W^*), \W -\W^*  \rangle + \frac{\gamma}{2} \|\W -\W^*\|^2  + \lambda\\
&=&\sum_{s\notin \Xi^*}\langle (\nabla \Psi(\W^*))_{s:}, (\W -\W^*)_{s:}  \rangle+ \frac{\gamma}{2} \|\W -\W^*\|^2+ \lambda \\
&\geq& - \sum_{s\notin \Xi^*}\|(\nabla \Psi(\W^*))_{s:}\|\cdot\|\W -\W^*\| + \frac{\gamma}{2} \|\W -\W^*\|^2+ \lambda\\
&\geq& - m \sqrt{2\lambda/\beta} \sqrt{\beta\lambda/(2m^2)} + \frac{\gamma}{2} \|\W -\W^*\|^2+ \lambda  \\
&=& \frac{\gamma}{2} \|\W -\W^*\|^2,
 \ea
 \end{eqnarray*}
where the first equality and the last inequality are from  \eqref{P-equivalent} and  $ \W\in {\mathbb N}(\W^*, \sqrt{\beta\lambda/(2m^2)})$, displaying \eqref{quadratic-growth-property} as well.
\end{proof}
\subsection{Proof Theorem \ref{P-first-order} }
\begin{proof} We note that $\Psi(\W)$ is a strongly smooth with a constant $\tau \|\V\|_2^2+\gamma$ and a strongly convex with a constant $\gamma$. Then the rest proof is similar to \cite[Theorem 1]{Zhou32021} and thus is omitted here.
\end{proof}

\subsection{Proof Theorem \ref{PGM-converge} }
\begin{proof} As  $\Psi(\W)$ is  strongly smooth with  $L_{\Psi}:=\tau \|\V\|_2^2+\gamma$, it follows from \cite[Theorem 5.1]{Beck2019} that
\be\label{W-descent}
\ba{cl}
 \Psi(\W^{\ell+1}) + \lambda \|\W^{\ell+1}\|_{0,2} -(\Psi(\W^{\ell}) + \lambda \|\W^{\ell}\|_{0,2} ) \leq -\frac{1/\beta-L_{\Psi}}{2}\|\W^{\ell+1}-\W^{\ell}\|^2,
\ea\ee
which by $\Psi(\W^{\ell}) + \lambda \|\W^{\ell}\|_{0,2}\geq 0$ means that sequence $\{\Psi(\W^{\ell}) + \lambda \|\W^{\ell}\|_{0,2}\}$ converges. Then taking the limit of both sides of \eqref{W-descent} yields  $\|\W^{\ell+1}-\W^{\ell}\|\to0$. Again by \cite[Theorem 5.1]{Beck2019},
 any accumulating point (say $\W^*$) of sequence $\{\W^{\ell}\}$ is a P-stationary point of problem \eqref{W-sub-problem31}. Moreover, we note that the  quadratic growth property in  Theorem \ref{P-first-order-local} states that any P-stationary point is a unique local minimizer and thus it is isolated. Finally,  three facts: \cite[Theorem 4.10]{More1983}, P-stationary point being isolated, and $\|\W^{\ell+1}-\W^{\ell}\|\to0$ yield that the whole sequence converges to a P-stationary point of \eqref{W-sub-problem31}.
\end{proof}

\subsection{Proof Theorem \ref{the-convergence}}
 \begin{proof}
a)   For $\U_h^{k+1}$, it is the solution to  \eqref{Int3-Uh} (or the first sub-problem of \eqref{Int2}), leading to
\be
\arraycolsep=1.4pt\def\arraystretch{1.25}
\ba{lcl} F(\CW^{k}, \{\CU_{\leq h-1}^{k},\U_h^{k+1}\}, \CV^{k})\leq F(\CW^{k}, \CU^{k}, \CV^{k}).\ea\ee
For $\W_{h}^{k+1}$,  it is a solution to \eqref{Int3-Wi-k}.  Then it follow from \eqref{def-prox} that
\begin{eqnarray*}
  \arraycolsep=1.4pt\def\arraystretch{1.25}
\begin{array}{lll}
    \W_{h}^{k+1} &\in&  {\rm Prox}_{\beta \lambda\|\cdot\|_{2,0}} ~( \W_h^{k+1}-\beta \Theta_h^{k+1} )\\
     &=&  {\rm argmin}_{\W}~\lambda \|\W\|_{2,0}+ \frac{1}{2\beta}\|\W-(\W_h^{k+1}-\beta \Theta_h^{k+1} )\|^2.
\end{array}
\end{eqnarray*}
where $\Theta_h^{k+1}:=\nabla_\W \Phi ( \W_{h}^{k+1}; \U_{h}^{k+1},\V_{h-1}^k)$ and $\Phi(\cdot)$ is defined by \eqref{Int3-Wi-obj}. This indicates that
\begin{eqnarray*}
  \arraycolsep=1.4pt\def\arraystretch{1.25}
\begin{array}{lll}
 \lambda \|\W_h^{k+1}\|_{2,0}+ \frac{1}{2\beta}\|\W_h^{k+1}-(\W_h^{k+1}-\beta\Theta_h^{k+1} )\|^2
 \leq   \lambda \|\W_h^{k} \|_{2,0}+ \frac{1}{2\beta}\|\W_h^{k}-(\W_h^{k+1}-\beta\Theta_h^{k+1})\|^2,
\end{array}
\end{eqnarray*}
which after simple manipulation yields
 \be\label{lemma-decreasing-W-eq}
\arraycolsep=1.4pt\def\arraystretch{1.25}
\ba{lcl}
 \langle \Theta_h^{k+1}, \W_{h}^{k+1}-\W_{h}^{k} \rangle + \lambda (\|\W_{h}^{k+1}\|_{2,0} - \|\W_{h}^{k}\|_{2,0}) \leq    \frac{1}{2\beta} \|  \W_{h}^{k+1}-\W_{h}^{k}\|^2.\ea\ee
Using the above fact enables to derive that
\be
\arraycolsep=1.4pt\def\arraystretch{1.25}
\ba{lcl}
&&F(\{\CW_{\leq h-1}^{k},\W_h^{k+1}\}, \{\CU_{\leq h-1}^{k},\U_h^{k+1}\}, \CV^{k})\\
&-& F(\{\CW_{\leq h-1}^{k},\W_h^{k}\}, \{\CU_{\leq h-1}^{k},\U_h^{k+1}\}, \CV^{k})\\
&=& \frac{\tau}{2} \|\U_{h}^{k+1}-\W_{h}^{k+1}\V_{h-1}^k\|^2+\frac{\gamma }{2}\|\W_{h}^{k+1}\|^2+ \lambda \|\W_{h}^{k+1}\|_{2,0}\\
&-&\frac{\tau}{2} \|\U_{h}^{k+1}-\W_{h}^k\V_{h-1}^k\|^2-\frac{\gamma }{2}\|\W_{h}^{k}\|^2- \lambda \|\W_{h}^{k}\|_{2,0}\\
&=& \langle \Theta_h^{k+1}, \W_{h}^{k+1}-\W_{h}^{k} \rangle+ \lambda (\|\W_{h}^{k+1}\|_{2,0} - \|\W_{h}^{k}\|_{2,0})\\
&-& \frac{\tau}{2} \| (\W_{h}^{k+1}-\W_{h}^{k})\V_{h-1}^k\|^2 - \frac{\gamma}{2} \|  \W_{h}^{k+1}-\W_{h}^{k}\|^2 \\
&\leq&  - \frac{\tau}{2} \| (\W_{h}^{k+1}-\W_{h}^{k})\V_{h-1}^k\|^2 - \frac{\gamma-1/\beta}{2} \|  \W_{h}^{k+1}-\W_{h}^{k}\|^2 \qquad (\text{by \eqref{lemma-decreasing-W-eq}})\\
&\leq&   - \frac{\gamma-1/\beta}{2} \|  \W_{h}^{k+1}-\W_{h}^{k}\|^2.\ea\ee
For $\V_{h-1}^{k+1}$, since it is a solution to \eqref{Int3-Vi}, the optimality condition is
 \be\label{opt-con-V}
\arraycolsep=1.4pt\def\arraystretch{1.25}
\ba{lcl}
\tau (\W_{h}^{k+1})^\top(\W_{h}^{k+1}\V_{h-1}^k-\U_{h}^{k+1})  + \pi (\V_{h-1}^{k+1}-(\U_{h-1}^{k})_{0/1}) = 0. \ea\ee
As a result,
\be\arraycolsep=1.4pt\def\arraystretch{1.25}
\ba{lcl}
&&F(\{\CW_{\leq h-1}^{k},\W_h^{k+1}\}, \{\CU_{\leq h-1}^{k},\U_h^{k+1}\}, \{\CV_{\leq h-2}^{k+1},\V_{h-1}^{k+1}\})-F(\{\CW_{\leq h-1}^{k},\W_h^{k+1}\}, \{\CU_{\leq h-1}^{k},\U_h^{k+1}\}, \{\CV_{\leq h-2}^{k},\V_{h-1}^{k}\})\\
&=&\frac{\tau}{2}\|\U_{h}^{k+1}-\W_{h}^{k+1}\V_{h-1}^{k+1}\|^2+\frac{\pi}{2}\|\V_{h-1}^{k+1}-(\U_{h-1}^{k})_{0/1}\|^2-\frac{\tau}{2}\|\U_{h}^{k+1}-\W_{h}^{k+1}\V_{h-1}^k\|^2+\frac{\pi}{2}\|\V_{h-1}^{k}-(\U_{h-1}^{k})_{0/1}\|^2\\
&=& \langle \tau (\W_{h}^{k+1})^\top(\W_{h}^{k+1}\V_{h-1}^k-\U_{h}^{k+1})  + \pi (\V_{h-1}^{k+1}-(\U_{h-1}^{k})_{0/1}), \V_{h-1}^{k+1}-\V_{h-1}^{k} \rangle\\
&-&\frac{\tau}{2}\| \W_{h}^{k+1}(\V_{h-1}^{k+1}-\V_{h-1}^{k} )\|^2-\frac{\pi}{2}\|\V_{h-1}^{k+1}-\V_{h-1}^{k} \|^2\\
&=&-\frac{\tau}{2}\| \W_{h}^{k+1}(\V_{h-1}^{k+1}-\V_{h-1}^{k} )\|^2-\frac{\pi}{2}\|\V_{h-1}^{k+1}-\V_{h-1}^{k} \|^2~~(\text{by \eqref{opt-con-V}})\\
&\leq& -\frac{\pi}{2}\|\V_{h-1}^{k+1}-\V_{h-1}^{k} \|^2.
\ea\ee
For $\U_{h-1}^{k+1}$, it is a solution to the fourth sub-problem of \eqref{Int2}, leading to
\be \arraycolsep=1.4pt\def\arraystretch{1.25}
\ba{lcl}
&&F(\{\CW_{\leq h-1}^{k},\W_h^{k+1}\}, \{\CU_{\leq h-2}^{k},\U_{h-1}^{k+1},\U_h^{k+1}\}, \{\CV_{\leq h-2}^{k},\V_{h-1}^{k}\})\\
&\leq&F(\{\CW_{\leq h-1}^{k},\W_h^{k+1}\}, \{\CU_{\leq h-2}^{k},\U_{h-1}^k,\U_h^{k+1}\}, \{\CV_{\leq h-2}^{k},\V_{h-1}^{k}\}).
\ea\ee
Same analysis can be applied to $\W_{h-1}^{k+1}, \V_{h-2}^{k+1}, \U_{h-2}^{k+1},\cdots,\W_{2}^{k+1}, \V_{1}^{k+1},\U_{1}^{k+1},\W_{1}^{k+1}$. Overall,  the above conditions allow us to conclude that
\be\label{decreasing-pro-0}
  \arraycolsep=1.4pt\def\arraystretch{1.25}
\ba{lcl}
&& F(\CW^{k+1},\CU^{k+1},\CV^{k+1})- F(\CW^k,\CU^k,\CV^k)\\
&=&F(\CW^{k+1},\CU^{k+1},\CV^{k+1})- F(\{\W_{1}^k,\CW_{\geq 2}^{k+1}\}, \CU^{k+1}, \CV^{k+1})\\
&+&F(\{\W_{1}^k,\CW_{\geq 2}^{k+1}\}, \CU^{k+1}, \CV^{k+1}) - F(\{\W_{1}^{k},\CW_{\geq 2}^{k+1}\}, \{\U_1^k,\CU_{\geq 2}^{k+1}\}, \CV^{k+1})\\
&+& \cdots\\
&+& F(\{\CW_{\leq h-1}^{k},\W_h^{k+1}\}, \{\CU_{\leq h-1}^{k},\U_h^{k+1}\}, \CV^k)-F(\CW^{k}, \{\CU_{\leq h-1}^{k},\U_h^{k+1}\}, \CV^k)\\
&+&F(\CW^{k}, \{\CU_{\leq h-1}^{k},\U_h^{k+1}\}, \CV^k) - F(\CW^{k},\CU^{k}, \CV^{k})\\
&\leq& - \frac{\gamma-1/\beta}{2}\sum_{i\in[h]} \|  \W_{i}^{k+1}-\W_{i}^{k}\|^2   -\frac{\pi}{2}\sum_{i\in[h-1]} \|\V_{i}^{k+1}-\V_{i}^{k} \|^2.
\ea\ee

b) It follows from \eqref{decreasing-pro-0}, $F\geq 0$, and $\gamma\geq1/\beta$ that sequence $\{F(\CW^k,\CU^k,\CV^k)\}$  is non-increasing and convergent.    Taking the limit of  both sides of \eqref{decreasing-pro} immediately display \eqref{gap-0}.

c) Let  $(\CW^*,\CU^*,\CV^*)$ be any accumulating point of sequence $\{(\CW^k,\CU^k,\CV^k)\}$, that is there exist a subsequence $\{(\CW^k,\CU^k,\CV^k)\}_{k\in\cal K}$ such that
\begin{eqnarray}\label{limit-k-*}
\begin{array}{lll}
\lim_{k(\in\cal K)\to\infty}(\CW^k,\CU^k,\CV^k)  = (\CW^*,\CU^*,\CV^*).
\end{array}
\end{eqnarray}
For   $(\CW^*,\CU^*,\CV^*)$, by letting $\b^*_s:=(\W_h^*\V_{h-1}^*)_{:s},$  $\Delta_s^*:=(\b^*_s)_{\max}-(\b^*_s)_{\phi_{\y_s}}$, and  $\epsilon_s^*\to(\Delta_s^*)^+$, we define  three index sets as
\begin{eqnarray*}
 \arraycolsep=1.4pt\def\arraystretch{1.25}
 \begin{array}{lll}
\Upsilon^*_>:=\left\{ s\in[N]:~  (\Delta_s^*)^2 > \frac{1}{N}\|\y_s-(\b^*_s)_\hd \|^2 \right\},\Upsilon^*_=:=\left\{ s\in[N]:~  (\Delta_s^*)^2 = \frac{1}{N}\|\y_s-(\b^*_s)_\hd \|^2 \right\},\\
\Upsilon^*_<:=\left\{ s\in[N]:~  (\Delta_s^*)^2 < \frac{1}{N}\|\y_s-(\b^*_s)_\hd \|^2 \right\}.
\end{array}
\end{eqnarray*}
 Similarly, we also define $\Upsilon^k_>, \Upsilon^k_=, \Upsilon^k_<$ for $(\CW^k,\CU^k,\CV^k)  $. Now we claim that
 \be\label{>-<-=}
\arraycolsep=1.4pt\def\arraystretch{1.25}
 \ba{lclll}
\Upsilon^*_> \subseteq \Upsilon^k_>, \qquad \Upsilon^*_< \subseteq \Upsilon^k_<, \qquad \Upsilon^*_= \supseteq \Upsilon^k_=,
\ea
\ee
for sufficiently large $k\in\cal K$. To show that, denote $\delta_s^*:=(\Delta_s^*)^2 - \frac{1}{N}\|\y_s-(\b^*_s)_\hd \|^2, s\in[N]$. Recall the assumption that every column of $\W_h^*\V_{h-1}^*$ only has one maximal entry, namely, $\b_s^*$ only has one maximal entry. This together with \eqref{limit-k-*} means that $\b_s^k:=(\W_h^k\V_{h-1}^k)_{:s}$  also has one maximal entry and $\phi_{\b_s^k} = \phi_{\b_s^*}$ for sufficiently large $k\in\cal K$, thereby leading to
\begin{eqnarray}\label{yb*-ybk}
 \arraycolsep=1.4pt\def\arraystretch{1.25}
 \begin{array}{lll}
 \|\y_s-(\b^*_s)_\hd \|^2 =  \|\y_s-(\b^k_s)_\hd \|^2.
\end{array}
\end{eqnarray}
 Moreover, condition \eqref{limit-k-*} also suffices to \begin{eqnarray}\label{Delta-limit}
 \arraycolsep=1.4pt\def\arraystretch{1.25}
 \begin{array}{lll}
 \lim_{k(\in{\cal K})\to\infty}\Delta_s^k:= \lim_{k(\in{\cal K})\to\infty}\left((\b^{k}_s)_{\max}-(\b^{k}_s)_{\phi_{\y_s}}\right)=\Delta_s^*,
 \end{array}
\end{eqnarray}
delivering  $|(\Delta_s^*)^2-(\Delta_s^k)^2| < \delta_s^*$ for sufficiently large $k\in\cal K$. Now, for any $s\in  \Upsilon^*_> $, it follows
\begin{eqnarray*}
 \arraycolsep=1.4pt\def\arraystretch{1.25}
 \begin{array}{lll}
(\Delta_s^k)^2 > (\Delta_s^*)^2 - \delta_s^* &=& (\Delta_s^*)^2 - \left((\Delta_s^*)^2 - \frac{1}{N}\|\y_s-(\b^*_s)_\hd \|^2\right)\\
& =& \frac{1}{N}\|\y_s-(\b^*_s)_\hd \|^2  = \frac{1}{N}\|\y_s-(\b^k_s)_\hd \|^2, ~~(\text{by \eqref{yb*-ybk}})
\end{array}
\end{eqnarray*}
which means  $s\in  \Upsilon^k_> $, resulting in $\Upsilon^*_> \subseteq \Upsilon^k_>$. Similar reasoning also allows us to show that $\Upsilon^*_< \subseteq \Upsilon^k_<$, and hence $\Upsilon^*_= \supseteq \Upsilon^k_=$ due to $(\Upsilon^*_>\cup \Upsilon^*_< \cup \Upsilon^*_=) =(\Upsilon^k_>\cup \Upsilon^k_< \cup \Upsilon^k_=)=[N]$. Overall, we prove \eqref{>-<-=}. Since $[N]$ contains finitely many entries, for sufficiently large $k\in\cal K$, there must exist a subset ${\cal J}:=\{k_1,k_2,\cdots\} \subseteq \cal K$ such that
\be\label{G-In2-W-J}
\arraycolsep=1.4pt\def\arraystretch{1.25}
 \ba{lclll}
&& \Upsilon^*_> \subseteq \Upsilon^{k_1}_> \equiv \Upsilon^{k_2}_> \equiv \cdots  =: \Upsilon_>, \Upsilon^*_< \subseteq \Upsilon^{k_1}_< \equiv \Upsilon^{k_2}_< \equiv \cdots=: \Upsilon_<,\Upsilon^*_= \supseteq \Upsilon^{k_1}_= \equiv \Upsilon^{k_2}_= \equiv \cdots=: \Upsilon_=,
\ea
\ee
where the inclusions are due to \eqref{>-<-=}. Then from \eqref{soultion-Uh-k1}, we have that, for any $k\in\cal J$,
\begin{eqnarray} \label{uk+1}
 \arraycolsep=1.4pt\def\arraystretch{1.25}
\forall s\in[N]:~~~(\U_h^{k+1})_{:s}=  \left\{\begin{array}{lll}
\b^{k}_s, & s\in \Upsilon_>,\\
\b^{k}_s~ \text{or}~\b^{k}_s+\epsilon_s^k \e_{\phi_{\y_s}}, ~~& s\in \Upsilon_=,\\
 \b^{k}_s+\epsilon_s^k \e_{{\phi_{\y_s}}}, & s\in \Upsilon_<.
\end{array}\right.
\end{eqnarray}
where $\epsilon_s^k\to (\Delta_s^k)^+$. It follows from $\epsilon \to c^+ \Leftrightarrow \epsilon= \lim_{t\to 0^+}  (c + t)$  that
\begin{eqnarray*}
 \arraycolsep=1.4pt\def\arraystretch{1.25} \begin{array}{llll}
\lim_{k(\in{\cal J})\to\infty}\epsilon_s^k &=& \lim_{k(\in{\cal J})\to\infty}  \lim_{t\to 0^+}  (\Delta_s^k + t)\\
&=& \lim_{k(\in{\cal J})\to\infty} \Delta_s^k+ \lim_{t\to 0^+}   t \\
&=& \Delta_s^*+ \lim_{t\to 0^+}   t&(\text{by \eqref{Delta-limit} and ${\cal J}\subseteq {\cal K}$})\\
& =&  \lim_{t\to 0^+} (\Delta_s^*+  t) = \epsilon_s^*. &(\text{by  $\epsilon_s^*\to (\Delta_s^*)^+$})\\
\end{array}
\end{eqnarray*}
Taking the limit of both sides of \eqref{uk+1} along with $k\in\cal J$ yields
\begin{eqnarray*}
 \arraycolsep=1.4pt\def\arraystretch{1.25}
\forall s\in[N]:~~~(\U_h^*)_{:s}=  \left\{\begin{array}{lll}
\b^*_s, &  s\in \Upsilon_>,\\
\b^*_s~ \text{or}~\b^*_s+\epsilon_s^* \e_{\phi_{\y_s}}, ~~&  s\in \Upsilon_=,\\
 \b^*_s+\epsilon_s^* \e_{{\phi_{\y_s}}}, &  s\in \Upsilon_<.
\end{array}\right.
\end{eqnarray*}
This together with \eqref{G-In2-W-J} enable us to obtain
\begin{eqnarray*}
 \arraycolsep=1.4pt\def\arraystretch{1.25}
\forall s\in[N]:~~~(\U_h^*)_{:s}=  \left\{\begin{array}{lll}
\b^*_s, &  s\in \Upsilon_>^*,\\
 \b^*_s, &  s\in \Upsilon_> \setminus \Upsilon_>^*,\\
\b^*_s~ \text{or}~\b^*_s+\epsilon_s^* \e_{\phi_{\y_s}}, ~&  s\in \Upsilon_=,\\
 \b^*_s+ \epsilon_s^* \e_{{\phi_{\y_s}}}, &  s\in \Upsilon_< \setminus \Upsilon_<^*,\\
 \b^*_s+\epsilon_s^* \e_{{\phi_{\y_s}}}, &  s\in \Upsilon_<^*,
\end{array}\right.=  \left\{\begin{array}{lll}
\b^*_s, &  s\in \Upsilon_>^*,\\
\b^*_s~ \text{or}~\b^*_s+\epsilon_s^* \e_{\phi_{\y_s}}, ~&  s\in \Upsilon_=^*,\\
 \b^*_s+\epsilon_s^* \e_{{\phi_{\y_s}}}, &  s\in \Upsilon_<^*,
\end{array}\right.
\end{eqnarray*}
which  recalling \eqref{soultion-Uh} and \eqref {Uh-problem-*} indicates
\be
\arraycolsep=1.4pt\def\arraystretch{1.25}
  \ba{lclll}
\U_{h}^*&\in&\operatorname{arg} \min\limits_{\U_h}& \frac{\tau}{2} \|\U_h-\W_{h}^*\V_{h-1}^*\|^2+\frac{1}{2N} \|\Y- (\U_h)_{\hd}\|^{2}.
\ea \nonumber
\ee
Similarly, we can also prove that
\be
\arraycolsep=1.4pt\def\arraystretch{1.25}
 \left\{\ba{lclll}
\V_{i}^*&=&\operatorname{arg} \min\limits_{\V_{i}}& \frac{\tau}{2}\|\U_{i+1}^*-\W_{i+1}^*\V_{i}\|^2+\frac{\pi}{2}\|\V_{i}-(\U_{i}^*)_{0/1}\|^2, &\forall i\in[h-1],\\
\U_i^*&\in&\operatorname{arg} \min\limits_{\U_i}& \frac{\tau}{2}\|\U_i-\W_i^*\V_{i-1}^*\|^2+ \frac{\pi}{2}\|\V_i^*-(\U_i)_{0/1}\|^2, &\forall i\in[h-1].\\
\ea\right.\nonumber
\ee
Finally, we prove that
\be\label{G-In3-W*}
\arraycolsep=1.4pt\def\arraystretch{1.25}
 \ba{lclll}
\W_i^*&\in& {\rm Prox}_{\beta \lambda\|\cdot\|_{2,0}}&\left(\W_i^*-\beta \nabla_\W \Phi ( \W_{i}^*; \U_{i}^*,\V_{i-1}^*)  \right),& \forall i\in[h].
\ea
\ee
Let $\Xi_i^{k+1}:=\{s\in[d_{i-1}]:(\W_i^{k+1})_{:s}\neq 0\}$ and $\Xi_i^*:=\{s\in[d_{i-1}]:(\W_i^*)_{:s}\neq 0\}$.  Since $[d_{i-1}]$ contains finitely many entries, for sufficiently large $k\in\cal K$, there must exist a subset $\cal J \subseteq \cal K$ such that
\be\label{G-In3-W-J}
\arraycolsep=1.4pt\def\arraystretch{1.25}
 \ba{lclll}
\Xi_i^{k+1}\equiv \Xi_i^*,~~\forall k\in\cal J.
\ea
\ee
It follows from \eqref{Int3-Wi-k}   that
\begin{eqnarray*}
\arraycolsep=1.4pt\def\arraystretch{1.25}
\ba{lcll}
\W_{i}^{k+1}&\in&  {\rm Prox}_{\beta \lambda\|\cdot\|_{2,0}} ( \W_i^{k+1}-\beta \Theta_i^{k+1} ),  \ea
\end{eqnarray*}
where $\Theta_i^{k+1}:=\nabla_\W \Phi ( \W_{i}^{k+1}; \U_{i}^{k+1},\V_{i-1}^k)$,  which by \eqref{P-equivalent} derives
 \be
  \arraycolsep=1.4pt\def\arraystretch{1.25}
\left\{ \ba{lll}
\|(\Theta_i^{k+1})_{s:}\||=0~\text{and}~ \|(\W_i^{k+1})_{s:}\|\geq
\sqrt{2\beta\lambda},~~ & \text{if} ~ \|(\W_i^{k+1})_{s:}\|\neq 0, \\
\|(\Theta_i^{k+1})_{s:}\|\leq \sqrt{2\lambda/\beta}, & \text{if} ~ \|(\W_i^{k+1})_{s:}\|= 0. \nonumber
 \ea \right.
 \ee
 This together with \eqref{G-In3-W-J} leads to, for any $k\in\cal J$,
  \be
  \arraycolsep=1.4pt\def\arraystretch{1.25}
\left\{ \ba{lll}
\|(\Theta_i^{k+1})_{s:}\|=0~\text{and}~ \|(\W_i^{k+1})_{s:}\|\geq
\sqrt{2\beta\lambda},~~ & \text{if} ~ s\in \Xi_i^*, \\
\|(\Theta_i^{k+1})_{s:}\|\leq \sqrt{2\lambda/\beta}, & \text{if} ~  s\notin \Xi_i^*.
 \ea \right. \nonumber
 \ee
 Now taking the limit of all items in the above formula along with  $k\in\cal J$ enables us to obtain
   \be
  \arraycolsep=1.4pt\def\arraystretch{1.25}
\left\{ \ba{lll}
\|(\nabla_\W \Phi ( \W_{i}^*; \U_{i}^*,\V_{i-1}^*))_{s:}\|=0~\text{and}~ \|(\W_i^*)_{s:}\|\geq
\sqrt{2\beta\lambda},~~ & \text{if} ~ s\in \Xi_i^*, \\
\|(\nabla_\W \Phi ( \W_{i}^*; \U_{i}^*,\V_{i-1}^*))_{s:}\|\leq \sqrt{2\lambda/\beta}, & \text{if} ~  s\notin \Xi_i^*.
 \ea \right.
 \ee
 This recalling  \eqref{P-equivalent} proves that \eqref{G-In3-W*}. Overall, we show that $(\CW^*,\CU^*,\CV^*)$ satisfies \eqref{G-In3}.
\end{proof}

\end{document}